\documentclass{article}

\usepackage{PRIMEarxiv}

\usepackage[utf8]{inputenc} 
\usepackage[T1]{fontenc}    
\usepackage{hyperref}       
\usepackage{url}            
\usepackage{booktabs}       
\usepackage{amsfonts}       
\usepackage{nicefrac}       
\usepackage{microtype}      
\usepackage{lipsum}
\usepackage{fancyhdr}       
\usepackage{tabularx}
\usepackage{ascii}
\usepackage{graphicx}       
\usepackage[dvipsnames]{xcolor}
\usepackage[inkscapelatex=false]{svg}
\graphicspath{{media/}}     
\usepackage{pgfplots}
\usepackage{lmodern}
\usepackage{subcaption}

\usepgfplotslibrary{fillbetween}
\usetikzlibrary{external}

\pgfplotsset{compat=1.16}
\usepgfplotslibrary{colormaps}

\pgfplotsset{
    colormap={shadesofred}{
        color(0) = (red!15);
        color(0.05) = (red!28);
        color(0.1) = (red!39);
        color(0.15) = (red!48);
        color(0.2) = (red!55);
        color(0.25) = (red!62);
        color(0.3) = (red!68);
        color(0.35) = (red!73);
        color(0.4) = (red!77);
        color(0.45) = (red);
        color(0.50) = (red);
        color(0.55) = (red!77);
        color(0.60) = (red!73);
        color(0.65) = (red!68);
        color(0.70) = (red!62);
        color(0.75) = (red!55);
        color(0.80) = (red!48);
        color(0.85) = (red!39);
        color(0.90) = (red!28);
        color(0.95) = (red!15);
    },
}

\usepackage{mathtools}

\title{Computational Life: How Well-formed, Self-replicating Programs Emerge from Simple Interaction}

\author{
  Blaise Agüera y Arcas\footnotemark[2]
  \and \bf
  Jyrki Alakuijala\footnotemark[2]
  \and \bf
  James Evans\footnotemark[3]
  \and \bf
  Ben Laurie\footnotemark[2]
  \and \bf
  Alexander Mordvintsev\footnotemark[2]
  \and \bf
  Eyvind Niklasson\footnotemark[2]
  \and \bf
  Ettore Randazzo\footnotemark[2]
  \and \bf
  Luca Versari\footnotemark[2] \\[1em]
  \footnotemark[2] Google, Paradigms of Intelligence Team and \footnotemark[3] The University of Chicago \\[1em]
  \texttt{\{blaisea, jyrki, benl, moralex, eyvind, etr, veluca\}@google.com} \\
  \texttt{jevans@uchicago.edu}
}

\begin{document}
\maketitle

\begin{abstract}
The fields of Origin of Life and Artificial Life both question what life is and how it emerges from a distinct set of ``pre-life'' dynamics. One common feature of most substrates where life emerges is a marked shift in dynamics when self-replication appears. While there are some hypotheses regarding how self-replicators arose in nature, we know very little about the general dynamics, computational principles, and necessary conditions for self-replicators to emerge. This is especially true on ``computational substrates'' where interactions involve logical, mathematical, or programming rules. In this paper we take a step towards understanding how self-replicators arise by studying several computational substrates based on various simple programming languages and machine instruction sets. We show that when random, non self-replicating programs are placed in an environment lacking any explicit fitness landscape, self-replicators tend to arise. We demonstrate how this occurs due to random interactions and self-modification, and can happen with and without background random mutations. We also show how increasingly complex dynamics continue to emerge following the rise of self-replicators. Finally, we show a counterexample of a minimalistic programming language where self-replicators are possible, but so far have not been observed to arise.

\end{abstract}

\keywords{Origins of Life \and Artificial Life \and Self-replication}

\section{Introduction}
The field of Origins of Life (OoL) has debated the definition of \textit{life} and the requirements and mechanisms for life to emerge since its inception~\cite{life10030020}. Different theories assign varying importance to the phenomena associated with living systems. Some consider the emergence of RNA as the major turning point~\cite{Gilbert1986}, while others focus on metabolism or chemical networks with autocatalytic properties~\cite{Wachtershauser1997-wu, kauffman2000investigations}. The question of what defines life and how it can emerge becomes necessarily more complex if we shift focus from ``life as it is'' to ``life as it could be'', the central question for the Artificial Life (ALife) community~\cite{Scharf2015-vx}. While searching for a general definition of \textit{life}, we observe a major change in dynamics coincident with the rise of self-replicators, which seems to apply regardless of substrate. Hence, we may use the appearance of self-replicators as a reasonable transition to distinguish \textit{pre-life} from \textit{life} dynamics~\cite{nowak2008}.

Many systems involve self-replication. RNA~\cite{Spiegelman1965-ej}, DNA, and associated polymerases are commonly accepted self-replicators. Autocatalytic networks are also widely considered self-replicators~\cite{Lancet2018-lm}. Self-replicators are also widespread in computer simulations by design. Most ALife experiments agents have predetermined methods of self-replication, but several experiments have also studied the dynamics of lower level and spontaneous self-replication. Famously, Cellular Automata (CA) were created to study self-replication and self-reproduction~\cite{neumann1966ca}. Self-replicating loops with CA have been extensively studied~\cite{LANGTON1984135,Sayama1999-ld,oros2007}. A recent extension of CA, Neural CA~\cite{mordvintsev2020growing}, can be trained to self-replicate patterns that robustly maintain interesting variation~\cite{Sinapayen2023}. Particle systems with suitable dynamical laws can also demonstrate self-replicating behaviors~\cite{Schmickl2016-nc}. Neural networks can be trained to output their own weights while performing auxiliary tasks~\cite{chang2018neural} and they can be trained to self-reproduce with meaningful variation in offspring~\cite{randazzo2021sr}. Finally, self-replicators can exist on computational substrates in the form of explicit programs that copy themselves, as in an assembly–like programming language~\cite{ray-approach-to-synthesis-1991,Ofria2004-zg}, or a LISP-based environment~\cite{Fontana1990AlgorithmicCA}, but this area of inquiry remains underexplored, and is the focus of this paper.

Much research on OoL and ALife focuses on the life period when self-replicators are already abundant. A central question during this period is: How do variation and complexity arise from simple self-replicators? Analyses often take the form of mathematical models and simulations~\cite{Takeuchi2012-ek}. In ALife, researchers often focus on selection for complex behaviors~\cite{ventrella1998genepool}, which may include interactions with other agents~\cite{miconi2008}. Simulations may include tens of thousands of parameters and complex virtual ecosystems~\cite{randazzo2023biomaker}, but they can rarely modify the means of self-replication beyond adapting the mutation rate. The two most notable exceptions use assembly-like languages as computational substrate. In Tierra~\cite{ray-approach-to-synthesis-1991}, simple assembly programs have no goals but are given time to execute their program and access and modify nearby memory. This causes them to self-replicate and manifest limited but interesting dynamics, including the rise of ``parasites'' that feed off other self-replicators. Avida~\cite{Ofria2004-zg} functions similarly: assembly-like programs are left running their code for a limited time. They can also self-replicate, this time by allocating new memory, writing their program in the new space, and then splitting. Avida adds a concept of fitness, since performing auxiliary computation increases a replicator's allotted execution time. Notably, both Tierra and Avida are seeded with a hand-crafted self-replicator, called the ``ancestor''. This puts them squarely into ``life'' dynamics, but still allows for modification of the self-replication mechanism.

But how does life begin? How do we get from a pre-life period devoid of self-replicators to one abundant with them? We know that several systems, initialized with randomly interacting primitives, can give rise to complex dynamics that result in selection under pre-life conditions~\cite{nowak2008}. The OoL field has extensively studied autocatalysis, chemical reactions where one of the reaction products is also a catalyst for the same reaction, as well as autocatalytic networks (or sets), groups of chemicals that form a closed loop of catalytic reactions~\cite{Hordijk2018-gn}. Autocatalysis appears fundamental to the emergence of life in the biological world. Moreover, autocatalytic networks arise inevitably with sufficiently distinctive catalysts in the prebiotic ``soup''~\cite{KAUFFMAN19861}. These have also been simulated in computational experiments~\cite{Lancet2018-lm, hutton2002, Fontana1990AlgorithmicCA, kruszewski2022, RASMUSSEN1990111, rasmussen1991matter}.

Fontana~\cite{Fontana1990AlgorithmicCA}, for example, simulates the emergence of autocatalytic networks on the computational substrate of the lambda calculus using LISP programs (or functions). Each element is a function that takes another function as input and outputs a new function. Thus, a graph of interactions can be constructed which, on occasion, gives rise to autocatalytic networks. Fontana also performed a ``Turing gas'' simulation, where a fixed number of programs randomly interact using the following ordered rule:
\begin{equation}
    f + g \longrightarrow f + g + f(g)
\end{equation}
Where $f$ and $g$ are some legal lambda calculus functions. To conserve a fixed number of programs, one of the three right-hand side programs was eliminated using rule-based criteria. Aside from autocatalytic networks, a very simple solution involves the emergence of an identity function $i$, yielding:
\begin{equation}
    i + i \longrightarrow 3i
\end{equation}
This program has strong fitness, and it was often observed that the entire gas converges to the identity. This can be considered a trivial \textit{replicator}, which in some experiments is explicitly disallowed by constraint.

In~\cite{kruszewski2022}, the authors use combinatorial logic to create an ``artificial chemistry'' founded upon basic building blocks. Their system preserves ``mass'' (operations neither create nor destroy building blocks) and results in simple autocatalytic behaviors, ever-growing structures, and periods of transient self-replicator emergence. While lambda calculus and combinatorial logic are related to programming languages in general, they represent distinct computational substrates. For example, creating RNA-like structures that can self-replicate arbitrary payloads may involve different approaches, depending on the substrate. Biology is steadily furthering insights regarding the conditions under which complex replicators such as RNA and DNA could have arisen and under which conditions. This question is underexplored for the general case, especially on computational substrates. Given recent advances in Artificial Intelligence, computational substrates could very well form the foundation for new forms of life and complex, evolving behavior.

In this paper we focus on computational substrates formed atop various programming languages. Here we highlight some of the most relevant previous investigations of the pre-life period on such substrates~\cite{RASMUSSEN1990111, rasmussen1991matter, PARGELLIS1996111}. In all of these investigations, and in ours as well, there is no explicit fitness function that drives complexification or self-replicators to arise. Nevertheless, complex dynamics happen due to the implicit competition for scarce resources (space, execution time, and sometimes energy).

In Coreworld~\cite{RASMUSSEN1990111, rasmussen1991matter}, the authors explore the substrate of programming languages with multiple programs executed in parallel and sharing the instruction (and data) tape. Programs consume a locally shared resource (energy) for executing each operation. The authors perform different runs where they observe complex dynamical systems resembling the pre-life period hypothesized in biology and observed in our experiments as well. In Coreworld, large structures appear alongside inescapable self-loops. Some simple self-replicators of two instructions (\texttt{MOV-SPL}) often take over. Interestingly, when the authors seed the environment with a functioning (more complex) self-replicator, self-replicators do not take over and eventually random mutations caused by their copy mechanism make them go extinct.

In~\cite{PARGELLIS1996111}, the author observes and quantifies the likelihood of self-replicators to arise with a given environment and programming language. The rise of self-replicators in this environment is however due to either random initialization or to random mutations of imperfect self-replicators (whom appearance is in turn due to random initialization). 

While the generation of self-replicators can indeed happen due to random initialization or solely due to mutations, in this paper we show that, for the majority of the configurations we explore, self-replicators arise mainly (or sometimes solely) due to self-modification. We show that initialising random programs in a variety of environments, all lacking an explicit fitness landscape, nevertheless give rise to self-replicators. We observe that self-replicators arise mostly due to self-modification and this can happen both with and without background random mutation. We primarily investigate extensions to the ``Brainfuck'' language~\cite{bfsource,enwiki:bf}, an esoteric language chosen for its simplicity, and show how self-replicators arise in a variety of related systems. We show experiments undertaken on an isolated system variant of the Turing gas in Fontana~\cite{Fontana1990AlgorithmicCA}, which we informally call ``primordial soup''. We then show how spatial extensions to the primordial soup cause self-replicators to arise with more interesting behaviors such as competition for space between different self-replicators. We also show how similar results are accomplished by extending the ``Forth''~\cite{Moore1970ForthA} programming language in different ways and in varying environments, as well as with real world instruction set of a Zilog Z80 8-bit microprocessor \cite{carr1980z80} emulator and with the Intel 8080 instruction set. Finally, we show a counterexample programming language, \texttt{SUBLEQ}~\cite{enwiki:subleq}, where we do not observe this transition from pre-life to life. We note that the shortest length of hand-crafted self-replicators in \texttt{SUBLEQ}-like substrates is significantly larger than what is observed in previous substrates.

\section{BFF: Extending Brainfuck}\label{sec:bff}
Brainfuck (BF) is an esoteric programming language widely known for its obscure minimalism. The original language consists of only eight basic commands, one data pointer, one instruction pointer, an input stream, and an output stream. Notably, the only mathematical operations are ``add one'' and ``subtract one'', making it onerous for humans to program with this language. We extend BF to operate in a self-contained universe where the data and instruction tapes are the same and programs modify themselves. We do so by replacing input and output streams with operations to copy from one head to another. The instruction pointer, the read and the write heads (\texttt{head0} and \texttt{head1}) all operate on the same \texttt{tape} (stored as one byte per pointer position, and initialized to zero). The instruction pointer starts at zero and reads the instruction at that position. Every instruction not listed below is a no-operation. The complete instruction set is as follows:

\begin{center}
\begin{tabular}{>{\tt}cc>{\tt}l}
< & &  head0 = head0 - 1 \\
> & & head0 = head0 + 1 \\
\{ & & head1 = head1 - 1 \\
\} & & head1 = head1 + 1 \\
- & & tape[head0] = tape[head0] - 1 \\
+ & & tape[head0] = tape[head0] + 1 \\
. & & tape[head1] = tape[head0] \\
, & & tape[head0] = tape[head1] \\
{[} & & if (tape[head0] == 0):{\rm~jump forwards to matching \texttt{]} command. } \\
] & & if (tape[head0] != 0):{\rm~jump backwards to matching \texttt{[} command. } \\
\end{tabular}
\end{center}

Parenthesis matching follows the usual rules, allowing nesting. If no matching parenthesis is found, the program terminates. The program also terminates after a fixed number of characters being read ($2^{13})$. Note that since instructions and data sit in the same place, they are encoded with a single byte. Therefore, out of the 256 possible characters, only 10 are valid instructions and 1 corresponds to the true ``zero'' used to exit loops. Any remaining values can be used to store data. By having neither input nor output streams, program strings can only interact with one another. None of our experiments will have any explicit fitness functions and programs will simply be left to execute code and overwrite themselves and neighbors based on their own instructions. As we will show, this is enough for self-replicators to emerge. Since the majority of investigations from this paper will be performed on a family of extended BF languages, we give this family of extensions the acronym ``BFF''.

\subsection{Primordial soup simulations}

The main kind of simulations we will use in this paper are a variant of the Turing gas from Fontana~\cite{Fontana1990AlgorithmicCA}. In this gas, a large number of programs (usually $2^{17}$) form a ``primordial soup''. Each program consists of 64 1-byte characters which are randomly initialized from a uniform distribution. In these simulations, no new programs are generated or removed -- change only occurs through self-modification or random background mutations. In each epoch, programs interact with one another by selecting random ordered pairs, concatenating them and executing the resulting code for a fixed number of steps or until the program ends. Because our programming languages read and write on the same tape, which is the program itself, these executions generally modify both initial programs. At the end, the programs are separated and returned to the soup for future consideration.

We can interpret the interaction between any two programs ($A$ and $B$) as an irreversible chemical reaction where order matters. This can be described as having a uniform distribution of catalysts $a$ and $b$ that interact with $A$ and $B$ as follows: 
\begin{equation}
   A + B \overset{a}\longrightarrow \operatorname{split}(\operatorname{exec}(AB)) = A' + B'
\end{equation}
\begin{equation}
   A + B \overset{b}\longrightarrow \operatorname{split}(\operatorname{exec}(BA)) = A'' + B''
\end{equation}
Where $\operatorname{exec}$ runs the concatenated programs and \textit{split} divides the result back into two 64 byte strings. As we will see, just this kind of interaction, \textit{even without background noise}, is sufficient to generate self-replicators. In their simplest form, we can see self-replicators as immediate autocatalytic reactions of a program $S$ and food $F$ that act as follows:
\begin{equation}
   S + F \overset{a}\longrightarrow \operatorname{split}(\operatorname{exec}(SF)) = 2 \cdot S
\end{equation}
This is because the self-replicator is unaffected by the code written in the other program and it gets repurposed as available real estate. Note that the behavior of the catalyst $b$ is undefined, but when the pool is full of self-replicators, this would result in either one of the two strings to self-replicate at random.

While useful for understanding operationally what occurs, we acknowledge that this framing has several limitations. First, it fails to account for autocatalysis that takes place over more than one step, which could occur for autocatalytic sets. Second, a self-replicator is generally much smaller than the full 64 byte window. If it copied itself with a specific offset different from 64, it may still count as a functional self-replication but it would fail to generate a perfect self-copy. This suggests that a more complete manner of inspection for the behavior of self-replicators would involve observing \textit{substrings}, but this is generally computationally intractable. We therefore will show a mixture of anecdotal evidence and graphs plotting summarizing complexity metrics.

\paragraph{Complexity metrics}
In this paper, we introduce a novel complexity metric we call ``high-order entropy''.
Theoretically, we define the \textbf{\textit{high-order entropy}} of a length $n$ string as
the difference between (1) its Shannon entropy (computed over individual tokens -- i.e. bytes) and (2) its ``normalized'' Kolmogorov complexity (i.e. its Kolmogorov complexity divided by $n$).

Intuitively, this complexity definition is meant to capture the amount of information that
can \textit{only} be explained by relations between different characters.

This metric shares similarities with sophistication~\cite{koppel1991almost,koppel1991learning,antunes2003sophistication} and effective complexity~\cite{gell2004effective}, because it attempts to ``factor out'' information in the string that comes from sampling i.i.d. variables. Nevertheless, we are not aware of methods to efficiently estimate these metrics. This led us to the construction of this new metric.

Properties of ``high-order entropy'' that justify its use as a complexity
metric and encode the above intuition include the following:

\begin{enumerate}
    \item\label{lowcplx} Given a sequence of $n$ i.i.d. characters, its expected high-order entropy converges to $0$ as $n$ grows to infinity.
    \item\label{highcplx} Given a sequence of $k$ i.i.d. characters with distribution $D$, the expected high-order entropy of the string formed by concatenating $n$ copies of those characters converges to the Shannon entropy of $D$ as $n$ grows to infinity.
\end{enumerate}

These two properties, combined, ensure that random noise will have no measurable complexity (property~\ref{lowcplx}),
while a soup obtained from many copies of the same string (as might arise from one ``taken over'' by a self-replicator) will
have substantial non-zero complexity.

While ``high-order entropy'' has nice theoretical properties, computing it requires knowing the Kolmogorov complexity of a string, which remains uncomputable~\cite{chaitin1974information}.
Moreover, insofar as in our experiments we use a pseudo-random number generator, the Kolmogorov complexity is by definition bounded by program size, which would not be the case if we were following true randomness.

Because of these challenges, in our experiments we approximated Kolmogorov complexity with the compressed size of the string achieved by a state-of-the-art text compressor\footnote{More precisely, the size achieved by compressing the string using the \texttt{brotli -q2} command that draws upon brotli v1.1.0.}. This follows the well-established practice in algorithmic information theory to use Lempel-Ziv-style compressors to approximate Kolmogorov complexity~(i.e.~\cite{cover1999elements,chen2000compression,horibe2003note}). This approximation results in a metric which is very fast to compute.

\paragraph{How self-replicators emerge: a case study}

In this section we are going to zoom in and analyze the dynamics of the state transition in one specific BFF run. We develop tools that facilitate this sort of analysis and help to pin-point the individual moment and location the replicator emerges, what happened before and after. To do so we use a technique inspired by radioactive tracers used in biology: we attach extra information tokens to all soup characters. Tokens are tuples of $({\tt epoch}, {\tt position}, {\tt char})$ packed into 64-bit integers. Whenever we put a new character in the soup (at initialization or because of a mutation), we create a new unique token. Copy operations {\tt.} and {\tt,} copy corresponding character tokens and displace tokens of overwritten characters. {\tt+} and {\tt-} operations only affect the {\tt char} part of the token, so the token origin can be traced after a balanced number of increments and decrements.

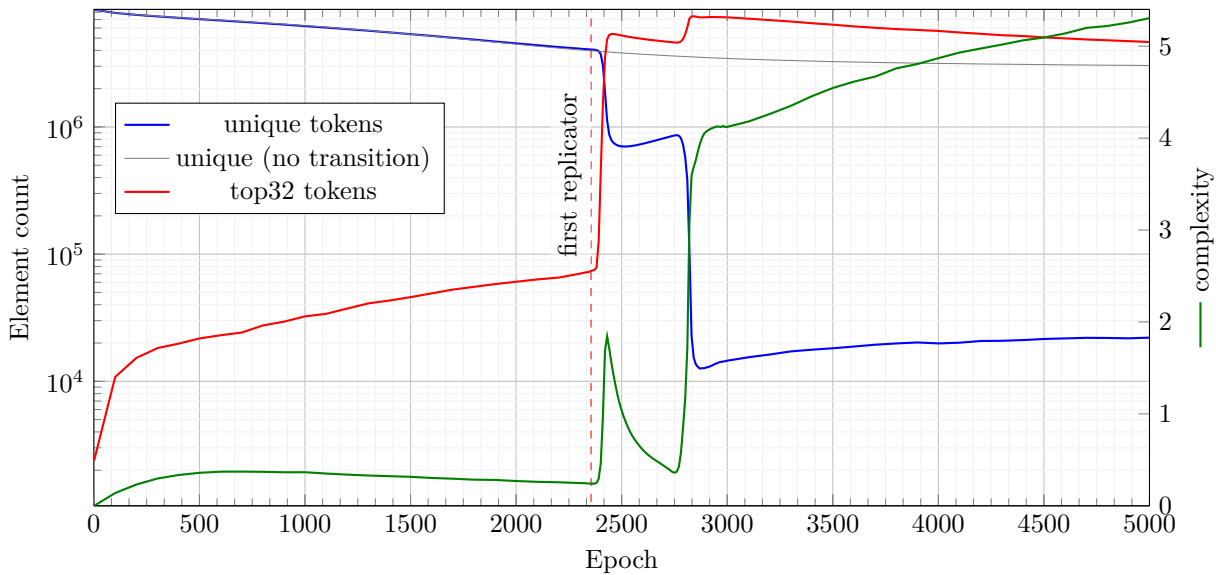
\begin{figure}
\pgfplotstableread{data0 unique_n top32_n complexity
1 8365708.0 2377.0 -0.000163
101 7882527.0 10819.0 0.138986
201 7614672.0 15251.0 0.231079
301 7401741.0 18250.0 0.296691
401 7218731.0 19775.0 0.335028
501 7042829.0 21748.0 0.358968
601 6877137.0 23029.0 0.371164
701 6715503.0 24195.0 0.371955
801 6551736.0 27506.0 0.369606
901 6382751.0 29451.0 0.364722
1001 6209224.0 32430.0 0.365076
1101 6045658.0 33963.0 0.350961
1201 5878611.0 37367.0 0.339205
1301 5713389.0 41035.0 0.329333
1401 5539040.0 43200.0 0.322227
1501 5365354.0 45935.0 0.315292
1601 5199743.0 49183.0 0.302798
1701 5024865.0 52669.0 0.294764
1801 4861503.0 55325.0 0.284292
1901 4694245.0 58150.0 0.281351
2001 4540273.0 60674.0 0.269446
2101 4392771.0 63298.0 0.260829
2201 4249674.0 65377.0 0.255887
2301 4127134.0 70170.0 0.247396
2311 4114761.0 70841.0 0.24707
2321 4102724.0 71405.0 0.246447
2331 4091670.0 71901.0 0.24464
2341 4080865.0 72184.0 0.243377
2351 4069728.0 73146.0 0.24137
2352 4068302.0 73210.0 0.241928
2353 4067199.0 73440.0 0.241503
2354 4066451.0 73603.0 0.24094
2355 4065308.0 73687.0 0.24179
2356 4064456.0 73812.0 0.241413
2357 4063015.0 73934.0 0.241683
2358 4061772.0 74086.0 0.242364
2359 4060313.0 74243.0 0.241832
2360 4059381.0 74247.0 0.241437
2361 4057970.0 74343.0 0.241836
2362 4057020.0 74343.0 0.241154
2363 4055915.0 74417.0 0.241651
2364 4054737.0 74533.0 0.240714
2365 4053412.0 74665.0 0.24093
2366 4052252.0 74696.0 0.240871
2367 4050896.0 74763.0 0.240341
2368 4049175.0 74871.0 0.240272
2369 4047527.0 75012.0 0.239542
2370 4045287.0 75095.0 0.240598
2371 4043489.0 75243.0 0.241402
2381 4020358.0 78530.0 0.249158
2391 3955322.0 126135.0 0.293789
2401 3728225.0 444803.0 0.462229
2411 3051369.0 1509489.0 0.966506
2421 1893296.0 3498127.0 1.685814
2431 1128391.0 4881398.0 1.843519
2441 873347.0 5298056.0 1.71527
2451 789712.0 5381979.0 1.561435
2461 750925.0 5379599.0 1.427298
2471 729457.0 5351339.0 1.306715
2481 715742.0 5316265.0 1.198971
2491 707898.0 5273961.0 1.10555
2501 703301.0 5234884.0 1.029952
2511 702090.0 5195740.0 0.960128
2521 702079.0 5157631.0 0.901641
2531 703405.0 5123721.0 0.847198
2541 706580.0 5090179.0 0.800113
2551 710554.0 5059741.0 0.759012
2561 714838.0 5030125.0 0.722102
2571 720214.0 5002311.0 0.690404
2581 726092.0 4974034.0 0.660877
2591 732082.0 4949141.0 0.632782
2601 738719.0 4923406.0 0.606339
2611 745720.0 4897595.0 0.584574
2621 752869.0 4873760.0 0.56262
2631 760263.0 4850877.0 0.543321
2641 767842.0 4828316.0 0.526271
2651 775104.0 4806817.0 0.508942
2661 783162.0 4786591.0 0.494282
2671 791577.0 4766228.0 0.478932
2681 799947.0 4744932.0 0.46506
2691 807975.0 4723996.0 0.448354
2701 815844.0 4705587.0 0.433814
2711 824702.0 4688030.0 0.416786
2721 833028.0 4671044.0 0.400713
2731 841763.0 4652626.0 0.384904
2741 850024.0 4633959.0 0.367374
2751 858306.0 4617286.0 0.360254
2761 861784.0 4607694.0 0.370579
2771 853276.0 4621698.0 0.424866
2781 815299.0 4698073.0 0.584087
2791 722171.0 4932780.0 0.8619
2801 578187.0 5353151.0 1.16951
2811 382119.0 6078973.0 1.718788
2821 103810.0 7119418.0 3.08987
2831 22804.0 7398822.0 3.595242
2841 15412.0 7409427.0 3.683469
2851 13628.0 7372953.0 3.759029
2861 12980.0 7317803.0 3.855959
2871 12587.0 7273493.0 3.941535
2881 12670.0 7274028.0 4.005048
2891 12683.0 7286936.0 4.050265
2901 12853.0 7293707.0 4.071874
2911 12943.0 7309189.0 4.087311
2921 13171.0 7314045.0 4.101686
2931 13388.0 7325575.0 4.10981
2941 13661.0 7318274.0 4.11931
2951 13852.0 7312903.0 4.126897
2961 14097.0 7305149.0 4.124252
2971 14214.0 7300855.0 4.122577
2981 14322.0 7287901.0 4.131279
2991 14402.0 7285819.0 4.12051
3001 14540.0 7280914.0 4.123148
3101 15463.0 7101682.0 4.180858
3201 16244.0 6922933.0 4.262331
3301 17190.0 6749774.0 4.350921
3401 17726.0 6542919.0 4.45615
3501 18198.0 6370067.0 4.546026
3601 18779.0 6175695.0 4.614991
3701 19404.0 6023997.0 4.66956
3801 19860.0 5885177.0 4.759122
3901 20207.0 5791842.0 4.8063
4001 19864.0 5686753.0 4.870472
4101 20126.0 5529481.0 4.931472
4201 20756.0 5401589.0 4.97486
4301 20823.0 5266018.0 5.017689
4401 21107.0 5187559.0 5.063466
4501 21524.0 5063661.0 5.094654
4601 21750.0 4971295.0 5.13939
4701 21972.0 4875429.0 5.198972
4801 21921.0 4803729.0 5.220189
4901 21785.0 4729963.0 5.257144
5001 22052.0 4653222.0 5.305795
}\dataZ
\pgfplotstableread{data1 unique1_n
1 8365375
101 7882580
201 7632282
301 7431042
401 7247341
501 7072150
601 6915188
701 6754264
801 6591680
901 6430701
1001 6265225
1101 6091042
1201 5906089
1301 5732553
1401 5546457
1501 5360818
1601 5180315
1701 4994675
1801 4818015
1901 4646583
2001 4476165
2101 4314547
2201 4172526
2301 4037792
2401 3922018
2501 3818654
2601 3729072
2701 3648731
2801 3577264
2901 3513441
3001 3458249
3101 3408532
3201 3366989
3301 3331522
3401 3295984
3501 3263846
3601 3240237
3701 3216973
3801 3197776
3901 3175577
4001 3159405
4101 3146623
4201 3128601
4301 3116463
4401 3104412
4501 3093001
4601 3080473
4701 3069578
4801 3058653
4901 3054690
5001 3043339
}\dataO

\begin{tikzpicture}

\begin{axis}[
    scale only axis,
    axis y line*=left,
    xlabel=Epoch,
    ylabel={Element count},
    width=0.85\textwidth,
    height=0.4\textwidth,
    grid=both,
    minor tick num=5,
    grid style={line width=.1pt, draw=gray!10},
    major grid style={line width=.2pt,draw=gray!50},
    xmin=0, xmax=5000,
    ymin=0, ymax=8.4e6,
    mark size=0,
    /pgf/number format/.cd,1000 sep={},
    ymode=log,
]
\addplot[color=blue,thick] table[y=unique_n] {\dataZ};
\label{unique}
\addplot[color=gray] table[y=unique1_n] {\dataO};
\label{uniquent}
\addplot[color=red,thick] table[y=top32_n] {\dataZ};
\label{top32}

\draw[dashed, red] (axis cs:2355,0) -- (axis cs:2355,8.4e6) node [pos=.8, above, sloped, color=black, fill=white, fill opacity=.5, text opacity=1] (TextNode) {first replicator};

\end{axis}

\begin{axis}[
    scale only axis,
    axis y line*=right,
    ylabel={\ref{complexity-plot-line} complexity},
    width=0.85\textwidth,
    height=0.4\textwidth,
    xmin=0, xmax=5000,
    ymin=0, ymax=5.4,
    mark size=0,
    legend style={at={(0.02,0.7)},anchor=west},
    xtick = {},
    xticklabels={},
]
\addlegendimage{/pgfplots/refstyle=unique}
\addlegendentry{unique tokens} 
\addlegendimage{/pgfplots/refstyle=uniquent}
\addlegendentry{unique (no transition)}
\addlegendimage{/pgfplots/refstyle=top32}
\addlegendentry{top32 tokens} 
\addplot[color=green!50!black,thick] table[y=complexity] {\dataZ}; 
\label{complexity-plot-line}
\end{axis}

\end{tikzpicture}
\caption{Tracer tokens and high-order entropy open a simple way to detect a state transition: we observe a rapid drop in the number of unique tokens, while the soup becomes dominated by a few most popular tokens. This is aligned with a state transition in complexity. Note that this particular state transition happened in two steps because of the ``zero-poisoning'' period (see Figure~\ref{fig:history}).}
\label{fig:tracer_log}
\end{figure}

The investigated simulation has a soup of $2^{17}$ tapes of $64$ characters, which gives $2^{23}=8 {\rm M}$ unique tokens at initialization. Token analysis also opens a simple way to detect a state transition (Figure~\ref{fig:tracer_log}).
Without a state transition, the number of unique tokens in the soup gradually decreases until it stabilizes at around $3{\rm M}$ unique tokens, when mutations and accidental replications counterbalance each other (Figure~\ref{fig:tracer_log}, black line). The state transition causes a sudden drop in the number of unique tokens and the soup becomes dominated by a few token ids. We observe that the token and complexity state transitions are perfectly aligned. This further corroborates the usefulness of using high-order entropy to detect the rise of self-replicators.

\begin{figure}
\includegraphics[width=\textwidth]{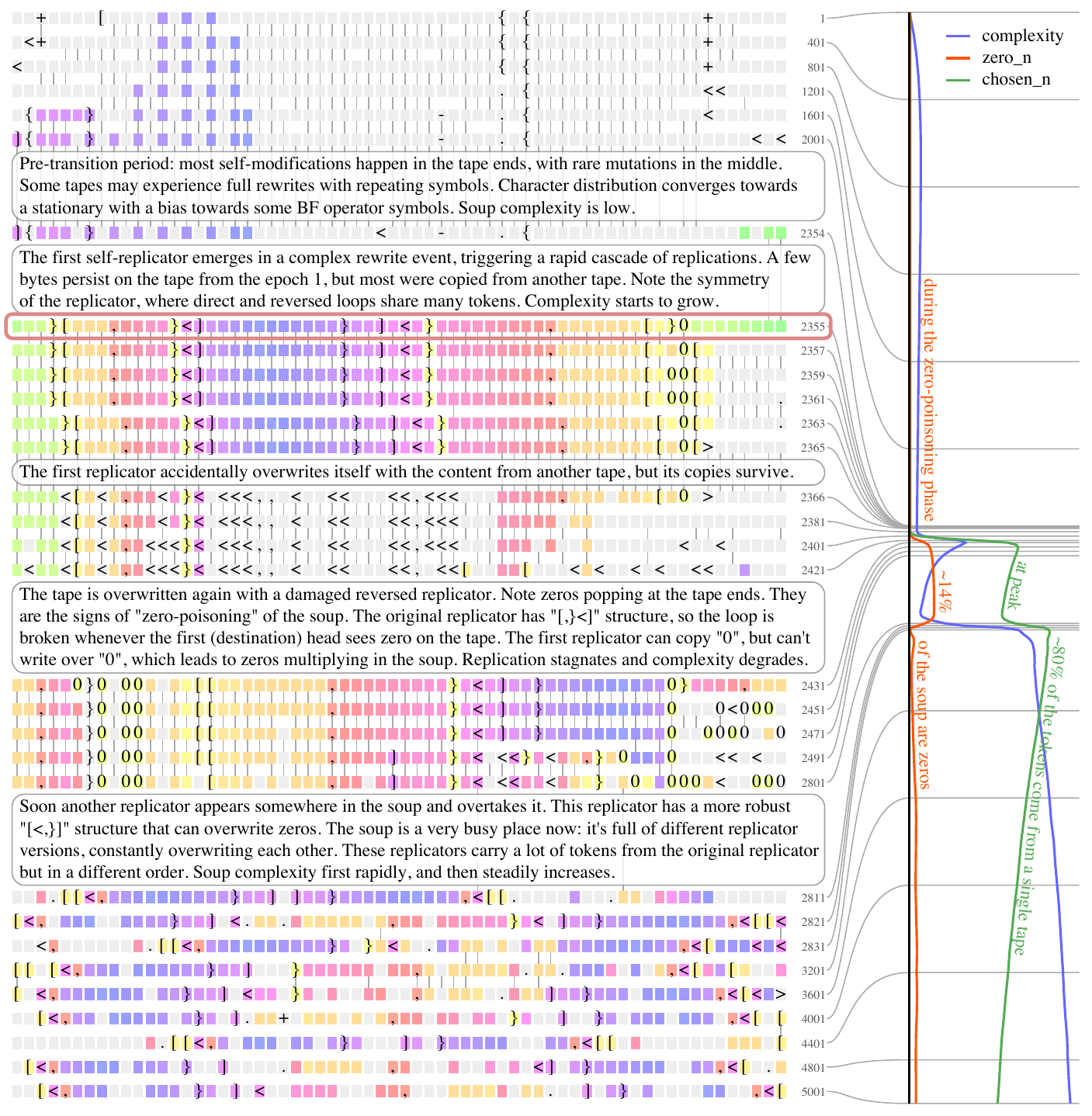}
\caption{The story of one state transition through the story of one tape. The left part of the figure shows snapshots of a single tape at different epochs. The selected tape is the one where the first (non-zero-tolerant) replicator emerges at epoch 2355 (see Figure~\ref{fig:first_replicator}). The emergence is followed by a ``zero-poisoning'' period, after which a new family of replicators takes over the soup. Only BF-code characters and zeros are printed. Lines connect consecutive character boxes if tokens match between the adjacent tape snapshots (which most often means that this place was not overwritten between snapshots). Colorful boxes correspond to the tokens that were present on the ``chosen tape'' at epoch 2355, this allows us to see where they came from and what happened to them later. The right part shows overall soup statistics: complexity (``high-order entropy''), number of zeros and number of tokens that match tokens of the ``chosen tape''.}
\label{fig:history}
\end{figure}

\begin{figure}
\begin{tikzpicture}
    \setlength\fboxsep{0.2pt}
    \node[anchor=south west,inner sep=0] (image) at (0,0) {\includegraphics[width=\textwidth]{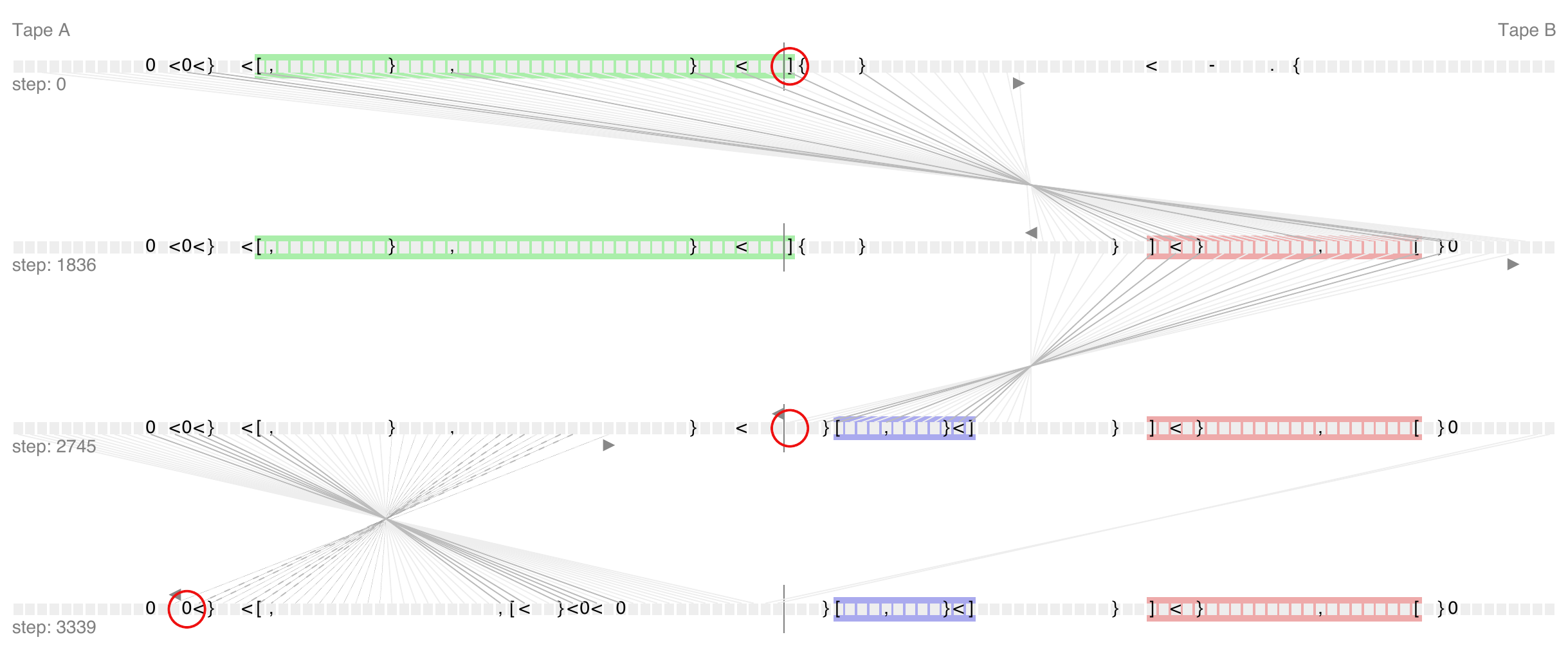}};
    \begin{scope}[x={(image.south east)},y={(image.north west)}]
        \node[text width=7cm,anchor=north west] at (0.03,0.87) {\scriptsize \colorbox[rgb]{0.66,0.93,0.66}{Pre-replicator loop} copies every second byte from tape A into B in reverse order. Note that this loop was incomplete without the tape B};
        
        \node[text width=7cm,anchor=north west] at (0.7,0.84) {\scriptsize \colorbox[rgb]{0.93,0.66,0.66}{Reversed replicator} is created in B};
        
        \node[text width=9cm,anchor=north west] at (0.03,0.56) {\scriptsize \colorbox[rgb]{0.66,0.93,0.66}{Pre-replicator} continues to copy the \colorbox[rgb]{0.93,0.66,0.66}{reversed replicator} into \colorbox[rgb]{0.66,0.66,0.93}{direct replicator} and overwrites “]”, thus breaking the first loop};

        \node[text width=10cm,anchor=north west] at (0.35,0.26) {\scriptsize \colorbox[rgb]{0.66,0.66,0.93}{Direct replicator} modifies tape A, but is terminated by “0” before it can overwrite the tape B. \textbf{The full replicator is complete in tape B now.}};
    \end{scope}
\end{tikzpicture}
\caption{Emergence of the self replicator at epoch 2354 of the case-study BFF run. Lines connect tokens that are copied from one tape to another.}
\label{fig:first_replicator}
\end{figure}

Tracing the origin of these tokens allowed us to exactly pinpoint the epoch and the tape whence the first replicator emerge. Figure~\ref{fig:history} gives a detailed overview and explanation of the observed dynamics and Figure~\ref{fig:first_replicator} highlights the precise events that cause the original self-replicator to arise.

\paragraph{Example self-replicator}

To show what happens during the execution of a self-replicator concatenated with another program, we extracted one self-replicator that appeared in our runs and sanitized it by clustering coding characters together and converting all non-coding characters to spaces. This operation does not change the behavior of a self-replicator. Then, we concatenated this program with a string full of zero-value bytes (this choice is arbitrary because the self-replicator ignores that context). Finally, we executed the concatenated program as shown in Figure~\ref{fig:example_selfrep}.

\begin{figure}
%
%

\begin{center}
{
\setlength{\fboxsep}{0pt}
\begin{tabularx}{\textwidth} { 
  >{\tt}r
  >{\centering\tiny\ttfamily\arraybackslash\fontseries{l}\selectfont}m{\textwidth - 2cm}}

1  &  \indent\rlap{\colorbox{blue!40}{\phantom{|}}}\rlap{\colorbox{red!70}{\phantom{|}}}\rlap{\colorbox{green}{\phantom{|}}}\fontseries{b}\selectfont[\fontseries{l}\selectfont\fontseries{b}\selectfont[\fontseries{l}\selectfont\fontseries{b}\selectfont\{\fontseries{l}\selectfont\fontseries{b}\selectfont.\fontseries{l}\selectfont\fontseries{b}\selectfont>\fontseries{l}\selectfont\fontseries{b}\selectfont]\fontseries{l}\selectfont\fontseries{b}\selectfont-\fontseries{l}\selectfont\fontseries{b}\selectfont]\fontseries{l}\selectfont\ \ \ \ \ \ \ \ \ \ \ \ \ \ \ \ \ \ \ \ \ \ \ \ \ \ \ \ \ \ \ \ \ \ \ \ \ \ \ \ \ \ \ \ \ \ \ \ \fontseries{b}\selectfont]\fontseries{l}\selectfont\fontseries{b}\selectfont-\fontseries{l}\selectfont\fontseries{b}\selectfont]\fontseries{l}\selectfont\fontseries{b}\selectfont>\fontseries{l}\selectfont\fontseries{b}\selectfont.\fontseries{l}\selectfont\fontseries{b}\selectfont\{\fontseries{l}\selectfont\fontseries{b}\selectfont[\fontseries{l}\selectfont\fontseries{b}\selectfont[\fontseries{l}\selectfont0000000000000000000000000000000000000000000000000000000000000000 \\
2  &  \indent\rlap{\colorbox{blue!40}{\phantom{|}}}\rlap{\colorbox{red!70}{\phantom{|}}}\fontseries{b}\selectfont[\fontseries{l}\selectfont\rlap{\colorbox{green}{\phantom{|}}}\fontseries{b}\selectfont[\fontseries{l}\selectfont\fontseries{b}\selectfont\{\fontseries{l}\selectfont\fontseries{b}\selectfont.\fontseries{l}\selectfont\fontseries{b}\selectfont>\fontseries{l}\selectfont\fontseries{b}\selectfont]\fontseries{l}\selectfont\fontseries{b}\selectfont-\fontseries{l}\selectfont\fontseries{b}\selectfont]\fontseries{l}\selectfont\ \ \ \ \ \ \ \ \ \ \ \ \ \ \ \ \ \ \ \ \ \ \ \ \ \ \ \ \ \ \ \ \ \ \ \ \ \ \ \ \ \ \ \ \ \ \ \ \fontseries{b}\selectfont]\fontseries{l}\selectfont\fontseries{b}\selectfont-\fontseries{l}\selectfont\fontseries{b}\selectfont]\fontseries{l}\selectfont\fontseries{b}\selectfont>\fontseries{l}\selectfont\fontseries{b}\selectfont.\fontseries{l}\selectfont\fontseries{b}\selectfont\{\fontseries{l}\selectfont\fontseries{b}\selectfont[\fontseries{l}\selectfont\fontseries{b}\selectfont[\fontseries{l}\selectfont0000000000000000000000000000000000000000000000000000000000000000 \\
3  &  \indent\rlap{\colorbox{blue!40}{\phantom{|}}}\rlap{\colorbox{red!70}{\phantom{|}}}\fontseries{b}\selectfont[\fontseries{l}\selectfont\fontseries{b}\selectfont[\fontseries{l}\selectfont\rlap{\colorbox{green}{\phantom{|}}}\fontseries{b}\selectfont\{\fontseries{l}\selectfont\fontseries{b}\selectfont.\fontseries{l}\selectfont\fontseries{b}\selectfont>\fontseries{l}\selectfont\fontseries{b}\selectfont]\fontseries{l}\selectfont\fontseries{b}\selectfont-\fontseries{l}\selectfont\fontseries{b}\selectfont]\fontseries{l}\selectfont\ \ \ \ \ \ \ \ \ \ \ \ \ \ \ \ \ \ \ \ \ \ \ \ \ \ \ \ \ \ \ \ \ \ \ \ \ \ \ \ \ \ \ \ \ \ \ \ \fontseries{b}\selectfont]\fontseries{l}\selectfont\fontseries{b}\selectfont-\fontseries{l}\selectfont\fontseries{b}\selectfont]\fontseries{l}\selectfont\fontseries{b}\selectfont>\fontseries{l}\selectfont\fontseries{b}\selectfont.\fontseries{l}\selectfont\fontseries{b}\selectfont\{\fontseries{l}\selectfont\fontseries{b}\selectfont[\fontseries{l}\selectfont\fontseries{b}\selectfont[\fontseries{l}\selectfont0000000000000000000000000000000000000000000000000000000000000000 \\
4  &  \indent\rlap{\colorbox{blue!40}{\phantom{|}}}\fontseries{b}\selectfont[\fontseries{l}\selectfont\fontseries{b}\selectfont[\fontseries{l}\selectfont\fontseries{b}\selectfont\{\fontseries{l}\selectfont\rlap{\colorbox{green}{\phantom{|}}}\fontseries{b}\selectfont.\fontseries{l}\selectfont\fontseries{b}\selectfont>\fontseries{l}\selectfont\fontseries{b}\selectfont]\fontseries{l}\selectfont\fontseries{b}\selectfont-\fontseries{l}\selectfont\fontseries{b}\selectfont]\fontseries{l}\selectfont\ \ \ \ \ \ \ \ \ \ \ \ \ \ \ \ \ \ \ \ \ \ \ \ \ \ \ \ \ \ \ \ \ \ \ \ \ \ \ \ \ \ \ \ \ \ \ \ \fontseries{b}\selectfont]\fontseries{l}\selectfont\fontseries{b}\selectfont-\fontseries{l}\selectfont\fontseries{b}\selectfont]\fontseries{l}\selectfont\fontseries{b}\selectfont>\fontseries{l}\selectfont\fontseries{b}\selectfont.\fontseries{l}\selectfont\fontseries{b}\selectfont\{\fontseries{l}\selectfont\fontseries{b}\selectfont[\fontseries{l}\selectfont\fontseries{b}\selectfont[\fontseries{l}\selectfont000000000000000000000000000000000000000000000000000000000000000\rlap{\colorbox{red!70}{\phantom{|}}}0\fontseries{l}\selectfont \\
5  &  \indent\rlap{\colorbox{blue!40}{\phantom{|}}}\fontseries{b}\selectfont[\fontseries{l}\selectfont\fontseries{b}\selectfont[\fontseries{l}\selectfont\fontseries{b}\selectfont\{\fontseries{l}\selectfont\fontseries{b}\selectfont.\fontseries{l}\selectfont\rlap{\colorbox{green}{\phantom{|}}}\fontseries{b}\selectfont>\fontseries{l}\selectfont\fontseries{b}\selectfont]\fontseries{l}\selectfont\fontseries{b}\selectfont-\fontseries{l}\selectfont\fontseries{b}\selectfont]\fontseries{l}\selectfont\ \ \ \ \ \ \ \ \ \ \ \ \ \ \ \ \ \ \ \ \ \ \ \ \ \ \ \ \ \ \ \ \ \ \ \ \ \ \ \ \ \ \ \ \ \ \ \ \fontseries{b}\selectfont]\fontseries{l}\selectfont\fontseries{b}\selectfont-\fontseries{l}\selectfont\fontseries{b}\selectfont]\fontseries{l}\selectfont\fontseries{b}\selectfont>\fontseries{l}\selectfont\fontseries{b}\selectfont.\fontseries{l}\selectfont\fontseries{b}\selectfont\{\fontseries{l}\selectfont\fontseries{b}\selectfont[\fontseries{l}\selectfont\fontseries{b}\selectfont[\fontseries{l}\selectfont000000000000000000000000000000000000000000000000000000000000000\rlap{\colorbox{red!70}{\phantom{|}}}\fontseries{b}\selectfont[\fontseries{l}\selectfont \\
6  &  \indent\fontseries{b}\selectfont[\fontseries{l}\selectfont\rlap{\colorbox{blue!40}{\phantom{|}}}\fontseries{b}\selectfont[\fontseries{l}\selectfont\fontseries{b}\selectfont\{\fontseries{l}\selectfont\fontseries{b}\selectfont.\fontseries{l}\selectfont\fontseries{b}\selectfont>\fontseries{l}\selectfont\rlap{\colorbox{green}{\phantom{|}}}\fontseries{b}\selectfont]\fontseries{l}\selectfont\fontseries{b}\selectfont-\fontseries{l}\selectfont\fontseries{b}\selectfont]\fontseries{l}\selectfont\ \ \ \ \ \ \ \ \ \ \ \ \ \ \ \ \ \ \ \ \ \ \ \ \ \ \ \ \ \ \ \ \ \ \ \ \ \ \ \ \ \ \ \ \ \ \ \ \fontseries{b}\selectfont]\fontseries{l}\selectfont\fontseries{b}\selectfont-\fontseries{l}\selectfont\fontseries{b}\selectfont]\fontseries{l}\selectfont\fontseries{b}\selectfont>\fontseries{l}\selectfont\fontseries{b}\selectfont.\fontseries{l}\selectfont\fontseries{b}\selectfont\{\fontseries{l}\selectfont\fontseries{b}\selectfont[\fontseries{l}\selectfont\fontseries{b}\selectfont[\fontseries{l}\selectfont000000000000000000000000000000000000000000000000000000000000000\rlap{\colorbox{red!70}{\phantom{|}}}\fontseries{b}\selectfont[\fontseries{l}\selectfont \\
7  &  \indent\fontseries{b}\selectfont[\fontseries{l}\selectfont\rlap{\colorbox{blue!40}{\phantom{|}}}\fontseries{b}\selectfont[\fontseries{l}\selectfont\rlap{\colorbox{green}{\phantom{|}}}\fontseries{b}\selectfont\{\fontseries{l}\selectfont\fontseries{b}\selectfont.\fontseries{l}\selectfont\fontseries{b}\selectfont>\fontseries{l}\selectfont\fontseries{b}\selectfont]\fontseries{l}\selectfont\fontseries{b}\selectfont-\fontseries{l}\selectfont\fontseries{b}\selectfont]\fontseries{l}\selectfont\ \ \ \ \ \ \ \ \ \ \ \ \ \ \ \ \ \ \ \ \ \ \ \ \ \ \ \ \ \ \ \ \ \ \ \ \ \ \ \ \ \ \ \ \ \ \ \ \fontseries{b}\selectfont]\fontseries{l}\selectfont\fontseries{b}\selectfont-\fontseries{l}\selectfont\fontseries{b}\selectfont]\fontseries{l}\selectfont\fontseries{b}\selectfont>\fontseries{l}\selectfont\fontseries{b}\selectfont.\fontseries{l}\selectfont\fontseries{b}\selectfont\{\fontseries{l}\selectfont\fontseries{b}\selectfont[\fontseries{l}\selectfont\fontseries{b}\selectfont[\fontseries{l}\selectfont000000000000000000000000000000000000000000000000000000000000000\rlap{\colorbox{red!70}{\phantom{|}}}\fontseries{b}\selectfont[\fontseries{l}\selectfont \\
8  &  \indent\fontseries{b}\selectfont[\fontseries{l}\selectfont\rlap{\colorbox{blue!40}{\phantom{|}}}\fontseries{b}\selectfont[\fontseries{l}\selectfont\fontseries{b}\selectfont\{\fontseries{l}\selectfont\rlap{\colorbox{green}{\phantom{|}}}\fontseries{b}\selectfont.\fontseries{l}\selectfont\fontseries{b}\selectfont>\fontseries{l}\selectfont\fontseries{b}\selectfont]\fontseries{l}\selectfont\fontseries{b}\selectfont-\fontseries{l}\selectfont\fontseries{b}\selectfont]\fontseries{l}\selectfont\ \ \ \ \ \ \ \ \ \ \ \ \ \ \ \ \ \ \ \ \ \ \ \ \ \ \ \ \ \ \ \ \ \ \ \ \ \ \ \ \ \ \ \ \ \ \ \ \fontseries{b}\selectfont]\fontseries{l}\selectfont\fontseries{b}\selectfont-\fontseries{l}\selectfont\fontseries{b}\selectfont]\fontseries{l}\selectfont\fontseries{b}\selectfont>\fontseries{l}\selectfont\fontseries{b}\selectfont.\fontseries{l}\selectfont\fontseries{b}\selectfont\{\fontseries{l}\selectfont\fontseries{b}\selectfont[\fontseries{l}\selectfont\fontseries{b}\selectfont[\fontseries{l}\selectfont00000000000000000000000000000000000000000000000000000000000000\rlap{\colorbox{red!70}{\phantom{|}}}0\fontseries{l}\selectfont\fontseries{b}\selectfont[\fontseries{l}\selectfont \\
9  &  \indent\fontseries{b}\selectfont[\fontseries{l}\selectfont\rlap{\colorbox{blue!40}{\phantom{|}}}\fontseries{b}\selectfont[\fontseries{l}\selectfont\fontseries{b}\selectfont\{\fontseries{l}\selectfont\fontseries{b}\selectfont.\fontseries{l}\selectfont\rlap{\colorbox{green}{\phantom{|}}}\fontseries{b}\selectfont>\fontseries{l}\selectfont\fontseries{b}\selectfont]\fontseries{l}\selectfont\fontseries{b}\selectfont-\fontseries{l}\selectfont\fontseries{b}\selectfont]\fontseries{l}\selectfont\ \ \ \ \ \ \ \ \ \ \ \ \ \ \ \ \ \ \ \ \ \ \ \ \ \ \ \ \ \ \ \ \ \ \ \ \ \ \ \ \ \ \ \ \ \ \ \ \fontseries{b}\selectfont]\fontseries{l}\selectfont\fontseries{b}\selectfont-\fontseries{l}\selectfont\fontseries{b}\selectfont]\fontseries{l}\selectfont\fontseries{b}\selectfont>\fontseries{l}\selectfont\fontseries{b}\selectfont.\fontseries{l}\selectfont\fontseries{b}\selectfont\{\fontseries{l}\selectfont\fontseries{b}\selectfont[\fontseries{l}\selectfont\fontseries{b}\selectfont[\fontseries{l}\selectfont00000000000000000000000000000000000000000000000000000000000000\rlap{\colorbox{red!70}{\phantom{|}}}\fontseries{b}\selectfont[\fontseries{l}\selectfont\fontseries{b}\selectfont[\fontseries{l}\selectfont \\
10  &  \indent\fontseries{b}\selectfont[\fontseries{l}\selectfont\fontseries{b}\selectfont[\fontseries{l}\selectfont\rlap{\colorbox{blue!40}{\phantom{|}}}\fontseries{b}\selectfont\{\fontseries{l}\selectfont\fontseries{b}\selectfont.\fontseries{l}\selectfont\fontseries{b}\selectfont>\fontseries{l}\selectfont\rlap{\colorbox{green}{\phantom{|}}}\fontseries{b}\selectfont]\fontseries{l}\selectfont\fontseries{b}\selectfont-\fontseries{l}\selectfont\fontseries{b}\selectfont]\fontseries{l}\selectfont\ \ \ \ \ \ \ \ \ \ \ \ \ \ \ \ \ \ \ \ \ \ \ \ \ \ \ \ \ \ \ \ \ \ \ \ \ \ \ \ \ \ \ \ \ \ \ \ \fontseries{b}\selectfont]\fontseries{l}\selectfont\fontseries{b}\selectfont-\fontseries{l}\selectfont\fontseries{b}\selectfont]\fontseries{l}\selectfont\fontseries{b}\selectfont>\fontseries{l}\selectfont\fontseries{b}\selectfont.\fontseries{l}\selectfont\fontseries{b}\selectfont\{\fontseries{l}\selectfont\fontseries{b}\selectfont[\fontseries{l}\selectfont\fontseries{b}\selectfont[\fontseries{l}\selectfont00000000000000000000000000000000000000000000000000000000000000\rlap{\colorbox{red!70}{\phantom{|}}}\fontseries{b}\selectfont[\fontseries{l}\selectfont\fontseries{b}\selectfont[\fontseries{l}\selectfont \\
11  &  \indent\fontseries{b}\selectfont[\fontseries{l}\selectfont\fontseries{b}\selectfont[\fontseries{l}\selectfont\rlap{\colorbox{blue!40}{\phantom{|}}}\rlap{\colorbox{green}{\phantom{|}}}\fontseries{b}\selectfont\{\fontseries{l}\selectfont\fontseries{b}\selectfont.\fontseries{l}\selectfont\fontseries{b}\selectfont>\fontseries{l}\selectfont\fontseries{b}\selectfont]\fontseries{l}\selectfont\fontseries{b}\selectfont-\fontseries{l}\selectfont\fontseries{b}\selectfont]\fontseries{l}\selectfont\ \ \ \ \ \ \ \ \ \ \ \ \ \ \ \ \ \ \ \ \ \ \ \ \ \ \ \ \ \ \ \ \ \ \ \ \ \ \ \ \ \ \ \ \ \ \ \ \fontseries{b}\selectfont]\fontseries{l}\selectfont\fontseries{b}\selectfont-\fontseries{l}\selectfont\fontseries{b}\selectfont]\fontseries{l}\selectfont\fontseries{b}\selectfont>\fontseries{l}\selectfont\fontseries{b}\selectfont.\fontseries{l}\selectfont\fontseries{b}\selectfont\{\fontseries{l}\selectfont\fontseries{b}\selectfont[\fontseries{l}\selectfont\fontseries{b}\selectfont[\fontseries{l}\selectfont00000000000000000000000000000000000000000000000000000000000000\rlap{\colorbox{red!70}{\phantom{|}}}\fontseries{b}\selectfont[\fontseries{l}\selectfont\fontseries{b}\selectfont[\fontseries{l}\selectfont \\
12  &  \indent\fontseries{b}\selectfont[\fontseries{l}\selectfont\fontseries{b}\selectfont[\fontseries{l}\selectfont\rlap{\colorbox{blue!40}{\phantom{|}}}\fontseries{b}\selectfont\{\fontseries{l}\selectfont\rlap{\colorbox{green}{\phantom{|}}}\fontseries{b}\selectfont.\fontseries{l}\selectfont\fontseries{b}\selectfont>\fontseries{l}\selectfont\fontseries{b}\selectfont]\fontseries{l}\selectfont\fontseries{b}\selectfont-\fontseries{l}\selectfont\fontseries{b}\selectfont]\fontseries{l}\selectfont\ \ \ \ \ \ \ \ \ \ \ \ \ \ \ \ \ \ \ \ \ \ \ \ \ \ \ \ \ \ \ \ \ \ \ \ \ \ \ \ \ \ \ \ \ \ \ \ \fontseries{b}\selectfont]\fontseries{l}\selectfont\fontseries{b}\selectfont-\fontseries{l}\selectfont\fontseries{b}\selectfont]\fontseries{l}\selectfont\fontseries{b}\selectfont>\fontseries{l}\selectfont\fontseries{b}\selectfont.\fontseries{l}\selectfont\fontseries{b}\selectfont\{\fontseries{l}\selectfont\fontseries{b}\selectfont[\fontseries{l}\selectfont\fontseries{b}\selectfont[\fontseries{l}\selectfont0000000000000000000000000000000000000000000000000000000000000\rlap{\colorbox{red!70}{\phantom{|}}}0\fontseries{l}\selectfont\fontseries{b}\selectfont[\fontseries{l}\selectfont\fontseries{b}\selectfont[\fontseries{l}\selectfont \\
13  &  \indent\fontseries{b}\selectfont[\fontseries{l}\selectfont\fontseries{b}\selectfont[\fontseries{l}\selectfont\rlap{\colorbox{blue!40}{\phantom{|}}}\fontseries{b}\selectfont\{\fontseries{l}\selectfont\fontseries{b}\selectfont.\fontseries{l}\selectfont\rlap{\colorbox{green}{\phantom{|}}}\fontseries{b}\selectfont>\fontseries{l}\selectfont\fontseries{b}\selectfont]\fontseries{l}\selectfont\fontseries{b}\selectfont-\fontseries{l}\selectfont\fontseries{b}\selectfont]\fontseries{l}\selectfont\ \ \ \ \ \ \ \ \ \ \ \ \ \ \ \ \ \ \ \ \ \ \ \ \ \ \ \ \ \ \ \ \ \ \ \ \ \ \ \ \ \ \ \ \ \ \ \ \fontseries{b}\selectfont]\fontseries{l}\selectfont\fontseries{b}\selectfont-\fontseries{l}\selectfont\fontseries{b}\selectfont]\fontseries{l}\selectfont\fontseries{b}\selectfont>\fontseries{l}\selectfont\fontseries{b}\selectfont.\fontseries{l}\selectfont\fontseries{b}\selectfont\{\fontseries{l}\selectfont\fontseries{b}\selectfont[\fontseries{l}\selectfont\fontseries{b}\selectfont[\fontseries{l}\selectfont0000000000000000000000000000000000000000000000000000000000000\rlap{\colorbox{red!70}{\phantom{|}}}\fontseries{b}\selectfont\{\fontseries{l}\selectfont\fontseries{b}\selectfont[\fontseries{l}\selectfont\fontseries{b}\selectfont[\fontseries{l}\selectfont \\
14  &  \indent\fontseries{b}\selectfont[\fontseries{l}\selectfont\fontseries{b}\selectfont[\fontseries{l}\selectfont\fontseries{b}\selectfont\{\fontseries{l}\selectfont\rlap{\colorbox{blue!40}{\phantom{|}}}\fontseries{b}\selectfont.\fontseries{l}\selectfont\fontseries{b}\selectfont>\fontseries{l}\selectfont\rlap{\colorbox{green}{\phantom{|}}}\fontseries{b}\selectfont]\fontseries{l}\selectfont\fontseries{b}\selectfont-\fontseries{l}\selectfont\fontseries{b}\selectfont]\fontseries{l}\selectfont\ \ \ \ \ \ \ \ \ \ \ \ \ \ \ \ \ \ \ \ \ \ \ \ \ \ \ \ \ \ \ \ \ \ \ \ \ \ \ \ \ \ \ \ \ \ \ \ \fontseries{b}\selectfont]\fontseries{l}\selectfont\fontseries{b}\selectfont-\fontseries{l}\selectfont\fontseries{b}\selectfont]\fontseries{l}\selectfont\fontseries{b}\selectfont>\fontseries{l}\selectfont\fontseries{b}\selectfont.\fontseries{l}\selectfont\fontseries{b}\selectfont\{\fontseries{l}\selectfont\fontseries{b}\selectfont[\fontseries{l}\selectfont\fontseries{b}\selectfont[\fontseries{l}\selectfont0000000000000000000000000000000000000000000000000000000000000\rlap{\colorbox{red!70}{\phantom{|}}}\fontseries{b}\selectfont\{\fontseries{l}\selectfont\fontseries{b}\selectfont[\fontseries{l}\selectfont\fontseries{b}\selectfont[\fontseries{l}\selectfont \\
15  &  \indent\fontseries{b}\selectfont[\fontseries{l}\selectfont\fontseries{b}\selectfont[\fontseries{l}\selectfont\rlap{\colorbox{green}{\phantom{|}}}\fontseries{b}\selectfont\{\fontseries{l}\selectfont\rlap{\colorbox{blue!40}{\phantom{|}}}\fontseries{b}\selectfont.\fontseries{l}\selectfont\fontseries{b}\selectfont>\fontseries{l}\selectfont\fontseries{b}\selectfont]\fontseries{l}\selectfont\fontseries{b}\selectfont-\fontseries{l}\selectfont\fontseries{b}\selectfont]\fontseries{l}\selectfont\ \ \ \ \ \ \ \ \ \ \ \ \ \ \ \ \ \ \ \ \ \ \ \ \ \ \ \ \ \ \ \ \ \ \ \ \ \ \ \ \ \ \ \ \ \ \ \ \fontseries{b}\selectfont]\fontseries{l}\selectfont\fontseries{b}\selectfont-\fontseries{l}\selectfont\fontseries{b}\selectfont]\fontseries{l}\selectfont\fontseries{b}\selectfont>\fontseries{l}\selectfont\fontseries{b}\selectfont.\fontseries{l}\selectfont\fontseries{b}\selectfont\{\fontseries{l}\selectfont\fontseries{b}\selectfont[\fontseries{l}\selectfont\fontseries{b}\selectfont[\fontseries{l}\selectfont0000000000000000000000000000000000000000000000000000000000000\rlap{\colorbox{red!70}{\phantom{|}}}\fontseries{b}\selectfont\{\fontseries{l}\selectfont\fontseries{b}\selectfont[\fontseries{l}\selectfont\fontseries{b}\selectfont[\fontseries{l}\selectfont \\

  & \dots \\
  
255  &  \indent\fontseries{b}\selectfont[\fontseries{l}\selectfont\fontseries{b}\selectfont[\fontseries{l}\selectfont\rlap{\colorbox{green}{\phantom{|}}}\fontseries{b}\selectfont\{\fontseries{l}\selectfont\fontseries{b}\selectfont.\fontseries{l}\selectfont\fontseries{b}\selectfont>\fontseries{l}\selectfont\fontseries{b}\selectfont]\fontseries{l}\selectfont\fontseries{b}\selectfont-\fontseries{l}\selectfont\fontseries{b}\selectfont]\fontseries{l}\selectfont\ \ \ \ \ \ \ \ \ \ \ \ \ \ \ \ \ \ \ \ \ \ \ \ \ \ \ \ \ \ \ \ \ \ \ \ \ \ \ \ \ \ \ \ \ \ \ \ \fontseries{b}\selectfont]\fontseries{l}\selectfont\fontseries{b}\selectfont-\fontseries{l}\selectfont\fontseries{b}\selectfont]\fontseries{l}\selectfont\fontseries{b}\selectfont>\fontseries{l}\selectfont\fontseries{b}\selectfont.\fontseries{l}\selectfont\fontseries{b}\selectfont\{\fontseries{l}\selectfont\fontseries{b}\selectfont[\fontseries{l}\selectfont\rlap{\colorbox{blue!40}{\phantom{|}}}\fontseries{b}\selectfont[\fontseries{l}\selectfont0\rlap{\colorbox{red!70}{\phantom{|}}}\fontseries{b}\selectfont[\fontseries{l}\selectfont\fontseries{b}\selectfont\{\fontseries{l}\selectfont\fontseries{b}\selectfont.\fontseries{l}\selectfont\fontseries{b}\selectfont>\fontseries{l}\selectfont\fontseries{b}\selectfont]\fontseries{l}\selectfont\fontseries{b}\selectfont-\fontseries{l}\selectfont\fontseries{b}\selectfont]\fontseries{l}\selectfont\ \ \ \ \ \ \ \ \ \ \ \ \ \ \ \ \ \ \ \ \ \ \ \ \ \ \ \ \ \ \ \ \ \ \ \ \ \ \ \ \ \ \ \ \ \ \ \ \fontseries{b}\selectfont]\fontseries{l}\selectfont\fontseries{b}\selectfont-\fontseries{l}\selectfont\fontseries{b}\selectfont]\fontseries{l}\selectfont\fontseries{b}\selectfont>\fontseries{l}\selectfont\fontseries{b}\selectfont.\fontseries{l}\selectfont\fontseries{b}\selectfont\{\fontseries{l}\selectfont\fontseries{b}\selectfont[\fontseries{l}\selectfont\fontseries{b}\selectfont[\fontseries{l}\selectfont \\
256  &  \indent\fontseries{b}\selectfont[\fontseries{l}\selectfont\fontseries{b}\selectfont[\fontseries{l}\selectfont\fontseries{b}\selectfont\{\fontseries{l}\selectfont\rlap{\colorbox{green}{\phantom{|}}}\fontseries{b}\selectfont.\fontseries{l}\selectfont\fontseries{b}\selectfont>\fontseries{l}\selectfont\fontseries{b}\selectfont]\fontseries{l}\selectfont\fontseries{b}\selectfont-\fontseries{l}\selectfont\fontseries{b}\selectfont]\fontseries{l}\selectfont\ \ \ \ \ \ \ \ \ \ \ \ \ \ \ \ \ \ \ \ \ \ \ \ \ \ \ \ \ \ \ \ \ \ \ \ \ \ \ \ \ \ \ \ \ \ \ \ \fontseries{b}\selectfont]\fontseries{l}\selectfont\fontseries{b}\selectfont-\fontseries{l}\selectfont\fontseries{b}\selectfont]\fontseries{l}\selectfont\fontseries{b}\selectfont>\fontseries{l}\selectfont\fontseries{b}\selectfont.\fontseries{l}\selectfont\fontseries{b}\selectfont\{\fontseries{l}\selectfont\fontseries{b}\selectfont[\fontseries{l}\selectfont\rlap{\colorbox{blue!40}{\phantom{|}}}\fontseries{b}\selectfont[\fontseries{l}\selectfont\rlap{\colorbox{red!70}{\phantom{|}}}0\fontseries{l}\selectfont\fontseries{b}\selectfont[\fontseries{l}\selectfont\fontseries{b}\selectfont\{\fontseries{l}\selectfont\fontseries{b}\selectfont.\fontseries{l}\selectfont\fontseries{b}\selectfont>\fontseries{l}\selectfont\fontseries{b}\selectfont]\fontseries{l}\selectfont\fontseries{b}\selectfont-\fontseries{l}\selectfont\fontseries{b}\selectfont]\fontseries{l}\selectfont\ \ \ \ \ \ \ \ \ \ \ \ \ \ \ \ \ \ \ \ \ \ \ \ \ \ \ \ \ \ \ \ \ \ \ \ \ \ \ \ \ \ \ \ \ \ \ \ \fontseries{b}\selectfont]\fontseries{l}\selectfont\fontseries{b}\selectfont-\fontseries{l}\selectfont\fontseries{b}\selectfont]\fontseries{l}\selectfont\fontseries{b}\selectfont>\fontseries{l}\selectfont\fontseries{b}\selectfont.\fontseries{l}\selectfont\fontseries{b}\selectfont\{\fontseries{l}\selectfont\fontseries{b}\selectfont[\fontseries{l}\selectfont\fontseries{b}\selectfont[\fontseries{l}\selectfont \\
257  &  \indent\fontseries{b}\selectfont[\fontseries{l}\selectfont\fontseries{b}\selectfont[\fontseries{l}\selectfont\fontseries{b}\selectfont\{\fontseries{l}\selectfont\fontseries{b}\selectfont.\fontseries{l}\selectfont\rlap{\colorbox{green}{\phantom{|}}}\fontseries{b}\selectfont>\fontseries{l}\selectfont\fontseries{b}\selectfont]\fontseries{l}\selectfont\fontseries{b}\selectfont-\fontseries{l}\selectfont\fontseries{b}\selectfont]\fontseries{l}\selectfont\ \ \ \ \ \ \ \ \ \ \ \ \ \ \ \ \ \ \ \ \ \ \ \ \ \ \ \ \ \ \ \ \ \ \ \ \ \ \ \ \ \ \ \ \ \ \ \ \fontseries{b}\selectfont]\fontseries{l}\selectfont\fontseries{b}\selectfont-\fontseries{l}\selectfont\fontseries{b}\selectfont]\fontseries{l}\selectfont\fontseries{b}\selectfont>\fontseries{l}\selectfont\fontseries{b}\selectfont.\fontseries{l}\selectfont\fontseries{b}\selectfont\{\fontseries{l}\selectfont\fontseries{b}\selectfont[\fontseries{l}\selectfont\rlap{\colorbox{blue!40}{\phantom{|}}}\fontseries{b}\selectfont[\fontseries{l}\selectfont\rlap{\colorbox{red!70}{\phantom{|}}}\fontseries{b}\selectfont[\fontseries{l}\selectfont\fontseries{b}\selectfont[\fontseries{l}\selectfont\fontseries{b}\selectfont\{\fontseries{l}\selectfont\fontseries{b}\selectfont.\fontseries{l}\selectfont\fontseries{b}\selectfont>\fontseries{l}\selectfont\fontseries{b}\selectfont]\fontseries{l}\selectfont\fontseries{b}\selectfont-\fontseries{l}\selectfont\fontseries{b}\selectfont]\fontseries{l}\selectfont\ \ \ \ \ \ \ \ \ \ \ \ \ \ \ \ \ \ \ \ \ \ \ \ \ \ \ \ \ \ \ \ \ \ \ \ \ \ \ \ \ \ \ \ \ \ \ \ \fontseries{b}\selectfont]\fontseries{l}\selectfont\fontseries{b}\selectfont-\fontseries{l}\selectfont\fontseries{b}\selectfont]\fontseries{l}\selectfont\fontseries{b}\selectfont>\fontseries{l}\selectfont\fontseries{b}\selectfont.\fontseries{l}\selectfont\fontseries{b}\selectfont\{\fontseries{l}\selectfont\fontseries{b}\selectfont[\fontseries{l}\selectfont\fontseries{b}\selectfont[\fontseries{l}\selectfont \\
258  &  \indent\fontseries{b}\selectfont[\fontseries{l}\selectfont\fontseries{b}\selectfont[\fontseries{l}\selectfont\fontseries{b}\selectfont\{\fontseries{l}\selectfont\fontseries{b}\selectfont.\fontseries{l}\selectfont\fontseries{b}\selectfont>\fontseries{l}\selectfont\rlap{\colorbox{green}{\phantom{|}}}\fontseries{b}\selectfont]\fontseries{l}\selectfont\fontseries{b}\selectfont-\fontseries{l}\selectfont\fontseries{b}\selectfont]\fontseries{l}\selectfont\ \ \ \ \ \ \ \ \ \ \ \ \ \ \ \ \ \ \ \ \ \ \ \ \ \ \ \ \ \ \ \ \ \ \ \ \ \ \ \ \ \ \ \ \ \ \ \ \fontseries{b}\selectfont]\fontseries{l}\selectfont\fontseries{b}\selectfont-\fontseries{l}\selectfont\fontseries{b}\selectfont]\fontseries{l}\selectfont\fontseries{b}\selectfont>\fontseries{l}\selectfont\fontseries{b}\selectfont.\fontseries{l}\selectfont\fontseries{b}\selectfont\{\fontseries{l}\selectfont\fontseries{b}\selectfont[\fontseries{l}\selectfont\fontseries{b}\selectfont[\fontseries{l}\selectfont\rlap{\colorbox{blue!40}{\phantom{|}}}\rlap{\colorbox{red!70}{\phantom{|}}}\fontseries{b}\selectfont[\fontseries{l}\selectfont\fontseries{b}\selectfont[\fontseries{l}\selectfont\fontseries{b}\selectfont\{\fontseries{l}\selectfont\fontseries{b}\selectfont.\fontseries{l}\selectfont\fontseries{b}\selectfont>\fontseries{l}\selectfont\fontseries{b}\selectfont]\fontseries{l}\selectfont\fontseries{b}\selectfont-\fontseries{l}\selectfont\fontseries{b}\selectfont]\fontseries{l}\selectfont\ \ \ \ \ \ \ \ \ \ \ \ \ \ \ \ \ \ \ \ \ \ \ \ \ \ \ \ \ \ \ \ \ \ \ \ \ \ \ \ \ \ \ \ \ \ \ \ \fontseries{b}\selectfont]\fontseries{l}\selectfont\fontseries{b}\selectfont-\fontseries{l}\selectfont\fontseries{b}\selectfont]\fontseries{l}\selectfont\fontseries{b}\selectfont>\fontseries{l}\selectfont\fontseries{b}\selectfont.\fontseries{l}\selectfont\fontseries{b}\selectfont\{\fontseries{l}\selectfont\fontseries{b}\selectfont[\fontseries{l}\selectfont\fontseries{b}\selectfont[\fontseries{l}\selectfont \\

  & \dots \\
331  &  \indent\fontseries{b}\selectfont[\fontseries{l}\selectfont\fontseries{b}\selectfont[\fontseries{l}\selectfont\rlap{\colorbox{green}{\phantom{|}}}\fontseries{b}\selectfont\{\fontseries{l}\selectfont\fontseries{b}\selectfont.\fontseries{l}\selectfont\fontseries{b}\selectfont>\fontseries{l}\selectfont\fontseries{b}\selectfont]\fontseries{l}\selectfont\fontseries{b}\selectfont-\fontseries{l}\selectfont\fontseries{b}\selectfont]\fontseries{l}\selectfont\ \ \ \ \ \ \ \ \ \ \ \ \ \ \ \ \ \ \ \ \ \ \ \ \ \ \ \ \ \ \ \ \ \ \ \ \ \ \rlap{\colorbox{red!70}{\phantom{|}}}\ \fontseries{l}\selectfont\ \ \ \ \ \ \ \ \ \fontseries{b}\selectfont]\fontseries{l}\selectfont\fontseries{b}\selectfont-\fontseries{l}\selectfont\fontseries{b}\selectfont]\fontseries{l}\selectfont\fontseries{b}\selectfont>\fontseries{l}\selectfont\fontseries{b}\selectfont.\fontseries{l}\selectfont\fontseries{b}\selectfont\{\fontseries{l}\selectfont\fontseries{b}\selectfont[\fontseries{l}\selectfont\fontseries{b}\selectfont[\fontseries{l}\selectfont\fontseries{b}\selectfont[\fontseries{l}\selectfont\fontseries{b}\selectfont[\fontseries{l}\selectfont\fontseries{b}\selectfont\{\fontseries{l}\selectfont\fontseries{b}\selectfont.\fontseries{l}\selectfont\fontseries{b}\selectfont>\fontseries{l}\selectfont\fontseries{b}\selectfont]\fontseries{l}\selectfont\fontseries{b}\selectfont-\fontseries{l}\selectfont\fontseries{b}\selectfont]\fontseries{l}\selectfont\ \ \ \ \ \ \ \ \ \ \rlap{\colorbox{blue!40}{\phantom{|}}}\ \fontseries{l}\selectfont\ \ \ \ \ \ \ \ \ \ \ \ \ \ \ \ \ \ \ \ \ \ \ \ \ \ \ \ \ \ \ \ \ \ \ \ \ \fontseries{b}\selectfont]\fontseries{l}\selectfont\fontseries{b}\selectfont-\fontseries{l}\selectfont\fontseries{b}\selectfont]\fontseries{l}\selectfont\fontseries{b}\selectfont>\fontseries{l}\selectfont\fontseries{b}\selectfont.\fontseries{l}\selectfont\fontseries{b}\selectfont\{\fontseries{l}\selectfont\fontseries{b}\selectfont[\fontseries{l}\selectfont\fontseries{b}\selectfont[\fontseries{l}\selectfont \\
332  &  \indent\fontseries{b}\selectfont[\fontseries{l}\selectfont\fontseries{b}\selectfont[\fontseries{l}\selectfont\fontseries{b}\selectfont\{\fontseries{l}\selectfont\rlap{\colorbox{green}{\phantom{|}}}\fontseries{b}\selectfont.\fontseries{l}\selectfont\fontseries{b}\selectfont>\fontseries{l}\selectfont\fontseries{b}\selectfont]\fontseries{l}\selectfont\fontseries{b}\selectfont-\fontseries{l}\selectfont\fontseries{b}\selectfont]\fontseries{l}\selectfont\ \ \ \ \ \ \ \ \ \ \ \ \ \ \ \ \ \ \ \ \ \ \ \ \ \ \ \ \ \ \ \ \ \ \ \ \ \rlap{\colorbox{red!70}{\phantom{|}}}\ \fontseries{l}\selectfont\ \ \ \ \ \ \ \ \ \ \fontseries{b}\selectfont]\fontseries{l}\selectfont\fontseries{b}\selectfont-\fontseries{l}\selectfont\fontseries{b}\selectfont]\fontseries{l}\selectfont\fontseries{b}\selectfont>\fontseries{l}\selectfont\fontseries{b}\selectfont.\fontseries{l}\selectfont\fontseries{b}\selectfont\{\fontseries{l}\selectfont\fontseries{b}\selectfont[\fontseries{l}\selectfont\fontseries{b}\selectfont[\fontseries{l}\selectfont\fontseries{b}\selectfont[\fontseries{l}\selectfont\fontseries{b}\selectfont[\fontseries{l}\selectfont\fontseries{b}\selectfont\{\fontseries{l}\selectfont\fontseries{b}\selectfont.\fontseries{l}\selectfont\fontseries{b}\selectfont>\fontseries{l}\selectfont\fontseries{b}\selectfont]\fontseries{l}\selectfont\fontseries{b}\selectfont-\fontseries{l}\selectfont\fontseries{b}\selectfont]\fontseries{l}\selectfont\ \ \ \ \ \ \ \ \ \ \rlap{\colorbox{blue!40}{\phantom{|}}}\ \fontseries{l}\selectfont\ \ \ \ \ \ \ \ \ \ \ \ \ \ \ \ \ \ \ \ \ \ \ \ \ \ \ \ \ \ \ \ \ \ \ \ \ \fontseries{b}\selectfont]\fontseries{l}\selectfont\fontseries{b}\selectfont-\fontseries{l}\selectfont\fontseries{b}\selectfont]\fontseries{l}\selectfont\fontseries{b}\selectfont>\fontseries{l}\selectfont\fontseries{b}\selectfont.\fontseries{l}\selectfont\fontseries{b}\selectfont\{\fontseries{l}\selectfont\fontseries{b}\selectfont[\fontseries{l}\selectfont\fontseries{b}\selectfont[\fontseries{l}\selectfont \\
333  &  \indent\fontseries{b}\selectfont[\fontseries{l}\selectfont\fontseries{b}\selectfont[\fontseries{l}\selectfont\fontseries{b}\selectfont\{\fontseries{l}\selectfont\fontseries{b}\selectfont.\fontseries{l}\selectfont\rlap{\colorbox{green}{\phantom{|}}}\fontseries{b}\selectfont>\fontseries{l}\selectfont\fontseries{b}\selectfont]\fontseries{l}\selectfont\fontseries{b}\selectfont-\fontseries{l}\selectfont\fontseries{b}\selectfont]\fontseries{l}\selectfont\ \ \ \ \ \ \ \ \ \ \ \ \ \ \ \ \ \ \ \ \ \ \ \ \ \ \ \ \ \ \ \ \ \ \ \ \ \rlap{\colorbox{red!70}{\phantom{|}}}\ \fontseries{l}\selectfont\ \ \ \ \ \ \ \ \ \ \fontseries{b}\selectfont]\fontseries{l}\selectfont\fontseries{b}\selectfont-\fontseries{l}\selectfont\fontseries{b}\selectfont]\fontseries{l}\selectfont\fontseries{b}\selectfont>\fontseries{l}\selectfont\fontseries{b}\selectfont.\fontseries{l}\selectfont\fontseries{b}\selectfont\{\fontseries{l}\selectfont\fontseries{b}\selectfont[\fontseries{l}\selectfont\fontseries{b}\selectfont[\fontseries{l}\selectfont\fontseries{b}\selectfont[\fontseries{l}\selectfont\fontseries{b}\selectfont[\fontseries{l}\selectfont\fontseries{b}\selectfont\{\fontseries{l}\selectfont\fontseries{b}\selectfont.\fontseries{l}\selectfont\fontseries{b}\selectfont>\fontseries{l}\selectfont\fontseries{b}\selectfont]\fontseries{l}\selectfont\fontseries{b}\selectfont-\fontseries{l}\selectfont\fontseries{b}\selectfont]\fontseries{l}\selectfont\ \ \ \ \ \ \ \ \ \ \rlap{\colorbox{blue!40}{\phantom{|}}}\ \fontseries{l}\selectfont\ \ \ \ \ \ \ \ \ \ \ \ \ \ \ \ \ \ \ \ \ \ \ \ \ \ \ \ \ \ \ \ \ \ \ \ \ \fontseries{b}\selectfont]\fontseries{l}\selectfont\fontseries{b}\selectfont-\fontseries{l}\selectfont\fontseries{b}\selectfont]\fontseries{l}\selectfont\fontseries{b}\selectfont>\fontseries{l}\selectfont\fontseries{b}\selectfont.\fontseries{l}\selectfont\fontseries{b}\selectfont\{\fontseries{l}\selectfont\fontseries{b}\selectfont[\fontseries{l}\selectfont\fontseries{b}\selectfont[\fontseries{l}\selectfont \\
334  &  \indent\fontseries{b}\selectfont[\fontseries{l}\selectfont\fontseries{b}\selectfont[\fontseries{l}\selectfont\fontseries{b}\selectfont\{\fontseries{l}\selectfont\fontseries{b}\selectfont.\fontseries{l}\selectfont\fontseries{b}\selectfont>\fontseries{l}\selectfont\rlap{\colorbox{green}{\phantom{|}}}\fontseries{b}\selectfont]\fontseries{l}\selectfont\fontseries{b}\selectfont-\fontseries{l}\selectfont\fontseries{b}\selectfont]\fontseries{l}\selectfont\ \ \ \ \ \ \ \ \ \ \ \ \ \ \ \ \ \ \ \ \ \ \ \ \ \ \ \ \ \ \ \ \ \ \ \ \ \rlap{\colorbox{red!70}{\phantom{|}}}\ \fontseries{l}\selectfont\ \ \ \ \ \ \ \ \ \ \fontseries{b}\selectfont]\fontseries{l}\selectfont\fontseries{b}\selectfont-\fontseries{l}\selectfont\fontseries{b}\selectfont]\fontseries{l}\selectfont\fontseries{b}\selectfont>\fontseries{l}\selectfont\fontseries{b}\selectfont.\fontseries{l}\selectfont\fontseries{b}\selectfont\{\fontseries{l}\selectfont\fontseries{b}\selectfont[\fontseries{l}\selectfont\fontseries{b}\selectfont[\fontseries{l}\selectfont\fontseries{b}\selectfont[\fontseries{l}\selectfont\fontseries{b}\selectfont[\fontseries{l}\selectfont\fontseries{b}\selectfont\{\fontseries{l}\selectfont\fontseries{b}\selectfont.\fontseries{l}\selectfont\fontseries{b}\selectfont>\fontseries{l}\selectfont\fontseries{b}\selectfont]\fontseries{l}\selectfont\fontseries{b}\selectfont-\fontseries{l}\selectfont\fontseries{b}\selectfont]\fontseries{l}\selectfont\ \ \ \ \ \ \ \ \ \ \ \rlap{\colorbox{blue!40}{\phantom{|}}}\ \fontseries{l}\selectfont\ \ \ \ \ \ \ \ \ \ \ \ \ \ \ \ \ \ \ \ \ \ \ \ \ \ \ \ \ \ \ \ \ \ \ \ \fontseries{b}\selectfont]\fontseries{l}\selectfont\fontseries{b}\selectfont-\fontseries{l}\selectfont\fontseries{b}\selectfont]\fontseries{l}\selectfont\fontseries{b}\selectfont>\fontseries{l}\selectfont\fontseries{b}\selectfont.\fontseries{l}\selectfont\fontseries{b}\selectfont\{\fontseries{l}\selectfont\fontseries{b}\selectfont[\fontseries{l}\selectfont\fontseries{b}\selectfont[\fontseries{l}\selectfont \\

\end{tabularx}
}
\end{center}
\caption{Example execution for a functioning self-replicator. See video of this run at: \url{https://asciinema.org/a/nXW8NFxiUtHiNtteJwXAXraFa}}
\label{fig:example_selfrep}
\end{figure}

The green rectangle is the instruction pointer, while the blue and red rectangles are read and write heads, respectively. We can see how the instruction pointer loops infinitely while the write head moves left and the read head moves right. At each cycle, one instruction is copied. Since the original self-replicator is a palindrome, this results in the entire self-replicator being correctly copied to the adjacent tape, despite being copied in reverse.

\paragraph{Observed Evolution of Complexity}
Scoping beyond individual runs, we now show the observed evolution of complexity for 1000 different runs, all with the same hyperparameters used in the example run.

\newcommand{\plotcomplexity}[2] {
\addplot[smooth, thick, color=#2] table[col sep=comma, header=false, x index=0, y expr=\thisrowno{1} * 8 /(64 * \thisrowno{2})] {#1};
}

\begin{figure}[h]

\begin{tikzpicture}
\newcommand{\defquantile}[1] {
\addplot[draw=none, forget plot, name path=perc#1] table[col sep=comma, header=false, x index=0, y expr=1+\thisrowno{#1}] {runstats.csv};
}
\newcommand{\shadequantile}[3] {
\addplot [fill=red, opacity=#3, forget plot] fill between[of=perc#1 and perc#2];
}
\newcommand{\quantilelegend}[2]{
\addlegendimage{area legend, fill=red, opacity=#1, draw=none}\addlegendentry{$#2$}
}

\begin{axis}[
    xlabel=Epoch,
    ylabel=high-order entropy,
    colorbar horizontal,
    colorbar style={
        at={(0,-0.25)},
        anchor=south west,
        xticklabel = {$\pgfmathparse{\tick*5}\pgfmathprintnumber[fixed,precision=1]{\pgfmathresult}$\%},
        xtick style={draw=none},
    },
    colormap name=shadesofred,
    colorbar as palette,
    width=\textwidth,
    height=0.5\textwidth,
    x tick label style={
        /pgf/number format/.cd,
            fixed,
            1000 sep={},
            fixed zerofill,
            precision=0,
        /tikz/.cd
    },
    every x tick label/.append style={font=\small},
    scaled x ticks = false,
    ymode=log,
    ytick={1,...,8},
    minor ytick = {1,1.2,...,8},
    yticklabel = {
        \pgfmathparse{e^\tick-1}
        $\pgfmathprintnumber[fixed,precision=1]{\pgfmathresult}$
    },
    grid=both,
    minor tick num=5,
    grid style={line width=.1pt, draw=gray!10},
    major grid style={line width=.2pt,draw=gray!50},
    xmin=0,
    ymin=1,
    ymax=8,
    mark size=0,
    xmax=16384,
]
\defquantile{1}
\defquantile{2}
\defquantile{3}
\defquantile{4}
\defquantile{5}
\defquantile{6}
\defquantile{7}
\defquantile{8}
\defquantile{9}
\defquantile{10}
\defquantile{11}
\defquantile{12}
\defquantile{13}
\defquantile{14}
\defquantile{15}
\defquantile{16}
\defquantile{17}
\defquantile{18}
\defquantile{19}
\defquantile{20}
\defquantile{21}

\shadequantile{1}{21}{0.15}
\shadequantile{2}{20}{0.15}
\shadequantile{3}{19}{0.15}
\shadequantile{4}{18}{0.15}
\shadequantile{5}{17}{0.15}
\shadequantile{6}{16}{0.15}
\shadequantile{7}{15}{0.15}
\shadequantile{8}{14}{0.15}
\shadequantile{9}{13}{0.15}
\shadequantile{10}{12}{1.0}

\end{axis}
\end{tikzpicture}
\caption{Distribution of complexity over time across 1000 different runs, with $0.024\%$ mutation rate. Each shade of red represents a different quantile range. 40\% of runs show a state transition within 16k epochs.}
\label{fig:complexity_analysis}
\end{figure}
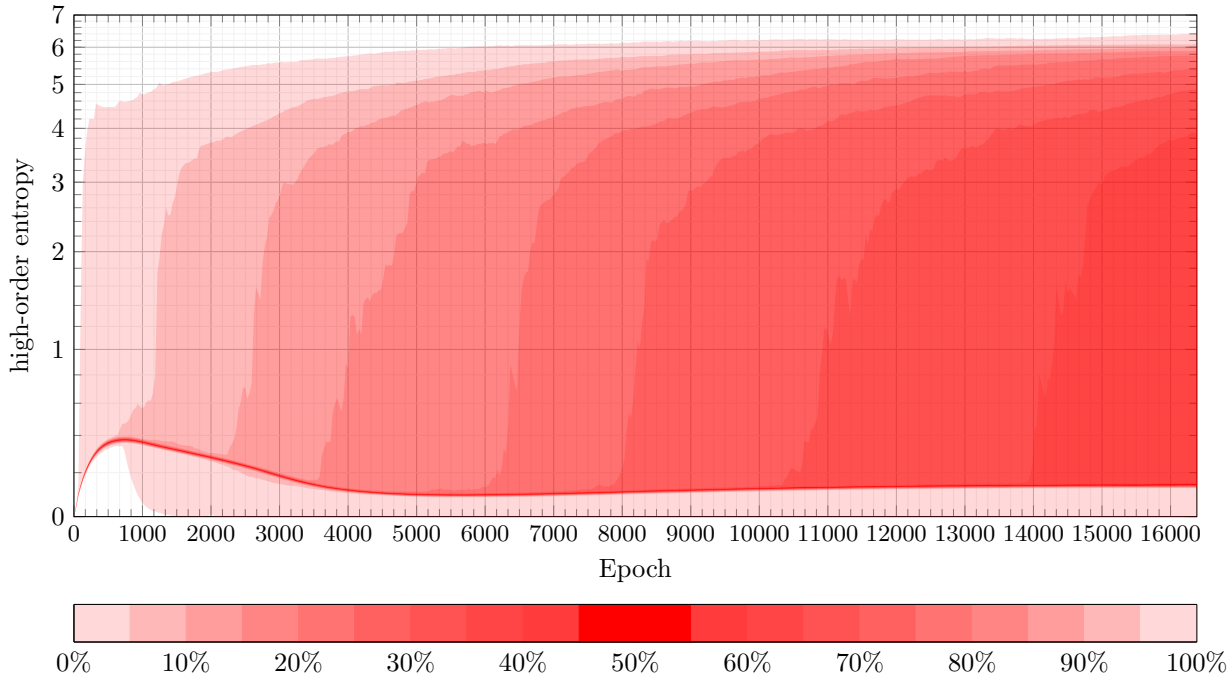

Figure~\ref{fig:complexity_analysis} shows how high-order entropy strangely, but consistently, increases in the first 1000 epochs, only to, on average, decrease again with a different distribution from the original (uniform) one. The different shades of red indicate how state transitions become increasingly more likely the more epochs pass. This reflects the appearance of stable self-replicators, which happens in 40\% of the runs within 16k epochs. Interestingly, some very lucky runs can have a state transition almost immediately.
    
\paragraph{Background noise ablation}

In the previous experiment, we focused on only one value for the background mutation rate. This does not show the average impact of random mutations for the appearance of self-replicators. For that, we perform an ablation study varying mutation rates.

\begin{figure}

\begin{tikzpicture}
\begin{axis}[
    colormap = {whitered}{color(0cm)  = (white);color(0.1cm)=(orange);color(0.4cm)=(red!10!orange); color(1cm) = (red)},
    colorbar,
    colorbar style={
        title=$ $,
        yticklabel style={
            /pgf/number format/.cd,
            fixed,
            precision=1,
            fixed zerofill,
        },
    },
    y tick label style={rotate=90,anchor=south},
    ylabel=mutations,
    ytick={0,1,2,3,4,5,6},
    yticklabels={$0\%$, $0.012\%$, $0.024\%$, $0.048\%$, $0.1\%$, $0.5\%$, $1\%$},
    xtick={0,1,2,3,4,5,6,7,8,9},
    xticklabels={
        $2000$,
        $4000$,
        $6000$,
        $8000$,
        $10000$,
        $12000$,
        $14000$,
        $16000$,
        N/A
    },
    nodes near coords={},
    width=0.9\textwidth,
    height=0.6\textwidth,
    title=Distribution of time to $\ge1$ complexity,
    axis on top,
    point meta min=0,
    point meta max=1,
    ymin=-0.5, ymax=6.5,
    xmin=-0.5, xmax=8.5,
    every node near coord/.append style={anchor=center},
    nodes near coords={\scriptsize\pgfmathprintnumber[fixed,precision=2]\pgfplotspointmeta},
]
\addplot [matrix plot*,point meta=explicit] table [x index=1, y index=0, meta index=2] {time_to_1_complexity.dat};
\end{axis}
\end{tikzpicture}

\caption{Distribution of the number of epochs needed for complexity to reach at least $1$ over $1000$ runs with different mutation rates. The ``N/A'' bucket includes all runs that did not get to $1$ complexity within 16k steps.}
\label{fig:mut_rates}
\end{figure}
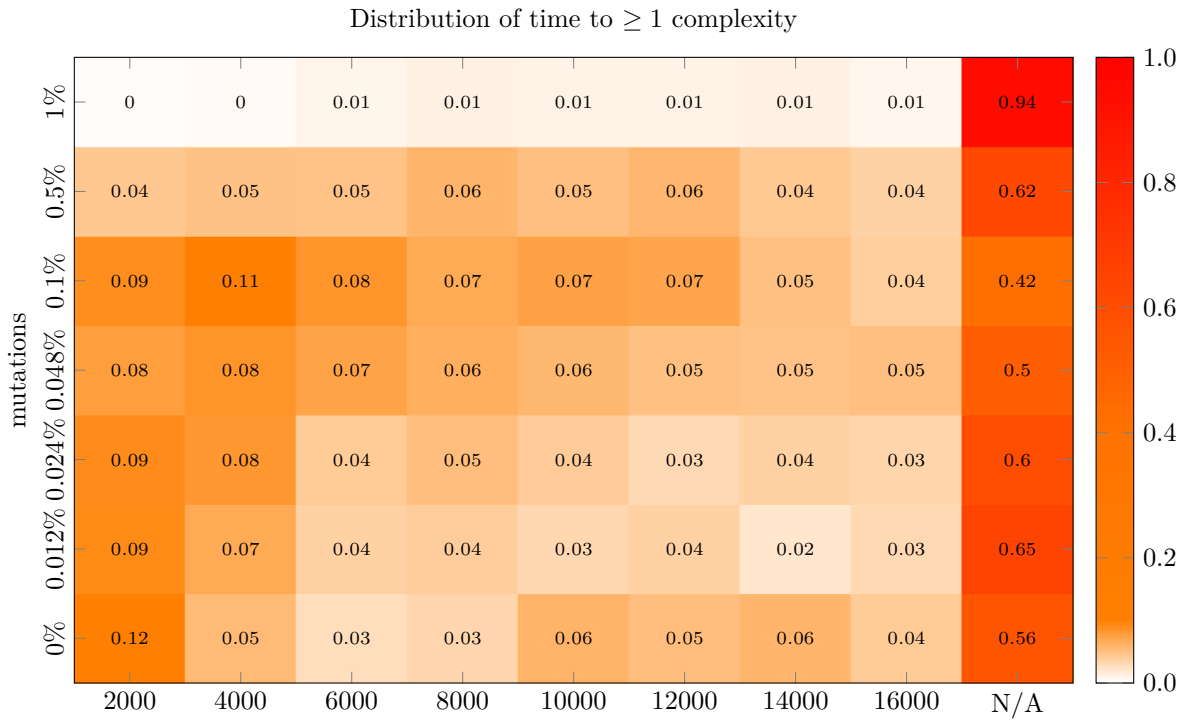

Figure~\ref{fig:mut_rates} shows a heat map for each mutation rate. Here as well, we observe how a state transition can occur at any point and there is a ~40\% chance of observing it within $16k$ steps with our default $0.024\%$ mutation rate. Generally, we observe that increasing the mutation rate speeds up the rise of self-replicators. While there is a theoretical limit of information that can be preserved with high mutation rates, as most self-replicators can be very small, we remain far from that limit within these experiments.

It is important to note that even without any background mutation, this state transition occurs with roughly the same frequency as our default case. This corroborates the fact that it is \textit{not} simply random bit-flipping that causes self-replicators to arise. Random flipping appears to speed up the process, but not having any would still cause a state transition. Moreover, the state transition is not localized only at the very beginning, suggesting that it is not simply random initialization that causes self-replicators to arise. The next experiment investigates this observation.

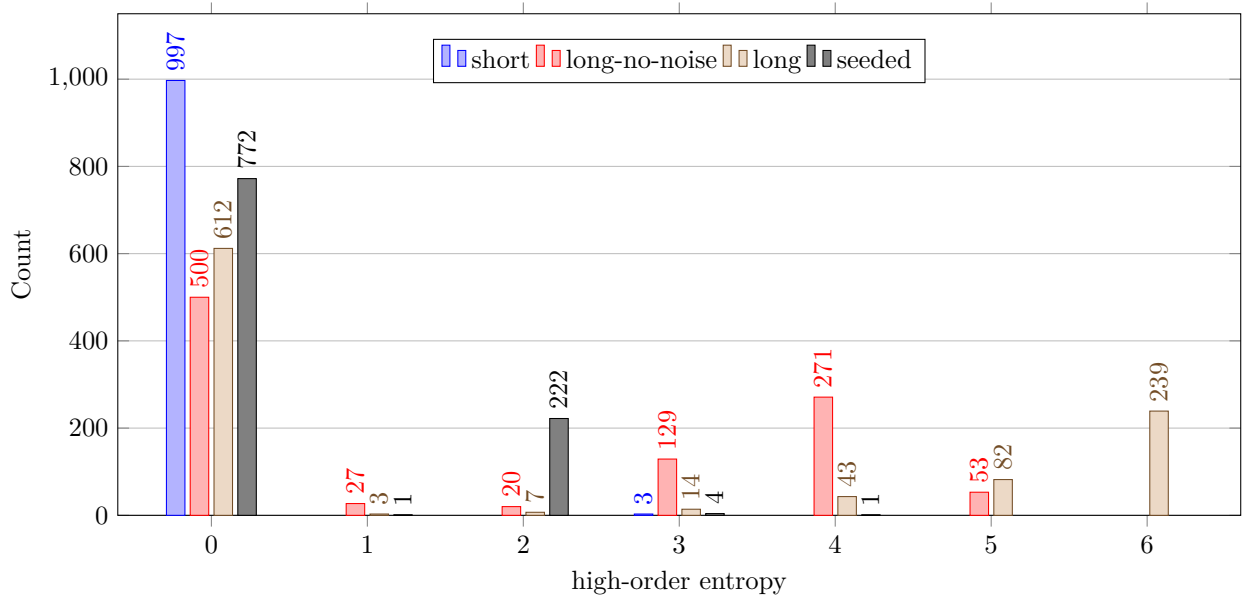
\begin{figure}
\begin{tikzpicture}
\begin{axis}[
    width = \textwidth,
    height = 0.5\textwidth,
	ylabel=Count,
	xtick={0,...,6},
	xlabel=high-order entropy,
	legend style={at={(0.5,0.95)},
	anchor=north,legend columns=-1},
	ybar, bar width = 7pt,
	nodes near coords,
	nodes near coords align={vertical},
	every node near coord/.append style={rotate=90, anchor=west},
	ymax = 1150,
	ymin = 0,
	ymajorgrids = true
]
\addplot coordinates {(0, 997) (3, 3)};
\addlegendentry{short}\label{histo-short}
\addplot coordinates {(0, 500) (1, 27) (2, 20) (3, 129) (4, 271) (5, 53)};
\addlegendentry{long-no-noise}\label{histo-long-no-noise}
\addplot coordinates {(0, 612) (1, 3) (2, 7) (3, 14) (4, 43) (5, 82) (6, 239)};
\addlegendentry{long}\label{histo-long}
\addplot coordinates {(0, 772) (1, 1) (2, 222) (3, 4) (4, 1)};
\addlegendentry{seeded}\label{histo-seeded}
\end{axis}
\end{tikzpicture}
\caption{Comparison of different kinds of runs, plotting the complexity of the soup at the end of the run. ``short'': random initialization run for 128 epochs; ``long'': random initialization run for 16k epochs; ``long-no-noise'': random initialization run for 16k epochs with no background noise and a fixed pattern of program interactions; `seeded': random initialization with an added functioning self-replicator in a random position, run for 128 epochs.}
\label{fig:init_comp}
\end{figure}

\paragraph{Comparison with random initialization}
How likely is it for a self-replicator to be present at initialization? This question is hard to answer precisely. As we discussed before, detecting a self-replicator is very hard, because it may not simply ``copy'' the entire string, or it may be part of a more complex autocatalytic set. The proxy we use is looking into high-order entropy which would show a state transition when self-replicators come to dominate. Nevertheless, it takes time for a self-replicator to take over an entire soup and during that period it can easily be destroyed. For example, 50\% of the time a single self-replicator will be the right-hand-side of the string and it may be irreversibly destroyed by the left-hand-side code. Morevover, random mutations may destroy it. To account for these concerns, we compare different kinds of runs in Figure~\ref{fig:init_comp}.

There, we perform 1000 runs for four kinds of experiments. The ``long'' label represents our usual run: random initialization and executing 16k epochs. We see how roughly 60\% of the runs fail to produce self-replicators. The ``short'' label represents random initialization and executing only 128 epochs. This is enough time for a self-replicator to take over if it is there at initialization, but note that it may be destroyed and that self-replicators may still appear due to self-modification and random mutations in this short time frame and not due to random initialization. However, we observe that a state transition is exceedingly rare, happening only 3 times out of 1000. For comparison, the ``seeded'' label indicates what would happen if we seeded the pool with one hand-written self-replicator (the one in Figure~\ref{fig:example_selfrep}) and let the system run for 128 epochs. There, 22\% of the time, a state transition happens, indicating that roughly $1/5$ of all self-replicators manage to take over. Finally, the difference in amounts of self-replicators arising between the ``short'' and the ``long'' runs may be due to the latter having much more entropy being added to the system, both in the form of background noise and the random interactions among different programs. While we already showed how runs with no background noise still generate plenty of self-replicators (Figure~\ref{fig:mut_rates}), we perform an extra comparison, ``long-no-noise'', where the mutation rate is zero \textit{and} we use a fixed sequence of shuffle patterns to determine the program pairings to not increase the overall entropy of the system. To our surprise, this variant is even more likely to result in a state transition (roughly 50\%) than the version with higher entropy shown in Figure~\ref{fig:mut_rates}.

With all of the experiments shown, we consider this sufficient evidence that self-replicators arise mostly due to self-modification and interaction among different programs and are not simply due to random initialization and random mutations.

\subsection{Spatial simulations}

\begin{figure}
\includegraphics[width=\textwidth]{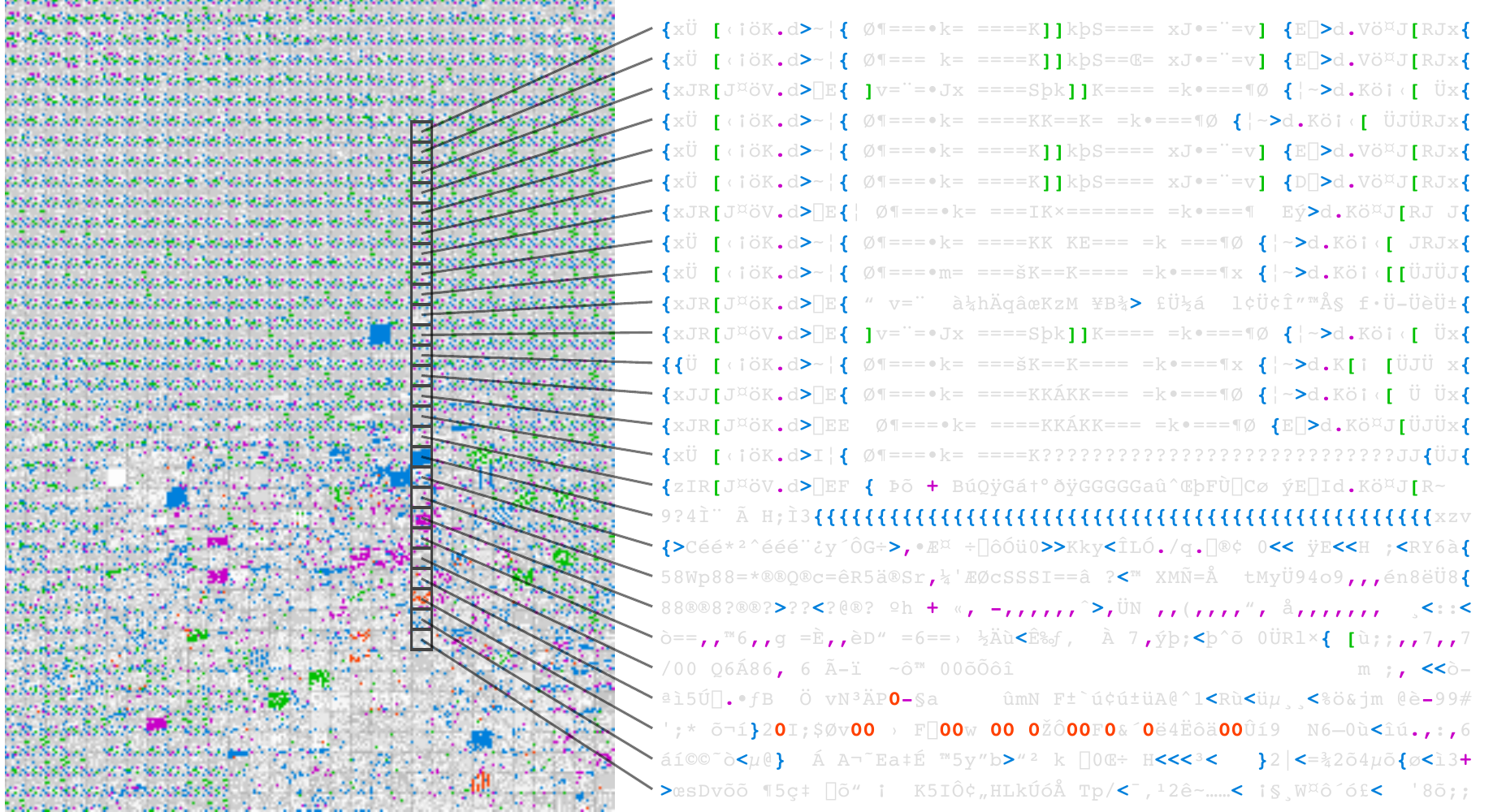}
\caption{A part of the 2D BFF soup in the process of state transition. Every 8x8 pixel square represents a single tape. Tapes are arranged in a 2d array and interactions only happen within a radius 2 neighbourhood of every cell. The small pre-transition area of the soup at bottom of the figure is being gradually overtaken by the wave of self-replicators.}
\label{fig:bff2d}
\end{figure}

The previous simulation of a primordial soup can be seen as an approximation of a 0-dimensional environment where all programs have a uniform chance of interacting with one another. We also experimented with imposing locality of communication in a BFF soup meant to replicate either a 1D and a 2D environment. While self-replicators emerge in all configurations, here we focus on the 2D environment.
More precisely, we set up a soup containing $32\,400$ BFF programs, arranged in a $240 \times 135$ grid.
We constrain programs to only be able to interact if $|x_0 - x_1| \le 2$ and $|y_0 - y_1| \le 2$, where $(x_0, y_0)$ and $(x_1, y_1)$ are the grid coordinates of the two programs; in words, two programs can only interact if their distance is at most $2$ along each coordinate.

At every epoch, we iterate through programs $P$ in a random order. For each program, we select one of its neighbours $N$, uniformly at random. Then, if neither $P$ or $N$ have been marked as taken in this epoch yet, we mark them as taken; finally, we apply the same execution rule as ``standard'' soup experiments on all the taken pairs, i.e. $\operatorname{split}(\operatorname{exec}(PN)) \longrightarrow P' + N'$. The resulting programs overwrite the original ones. Note that programs that do not execute still get mutated.

In the resulting simulation, self-replicators still emerge, as shown in Figure~\ref{fig:bff2d} and the accompanying YouTube video\footnote{\url{https://www.youtube.com/watch?v=07NoZwvgJ_M}}.
The main difference compared to the usual setup is given by the speed of propagation of self-replicators:
if all tapes are allowed to interact in a soup of size $n$, once a self-replicator emerges it typically takes over at least half of the soup in about
$\log n$ steps; on the other hand, in a 2D soup it takes a number of epochs that is proportional to the grid side lengths,
which is $\sqrt n$ for a square grid.

Because of this difference, 2D grid experiments are very helpful to visualize how self-replicators evolve and their behaviour.
It also provides a fertile ground for multiple variations of self-replicators to co-exist and compete with each other.

\subsection{Code for experiments}
The code for the experiments above can be found at \url{https://github.com/paradigms-of-intelligence/cubff}.
The code is written to support both CUDA and CPU execution; either can be chosen at compile time, but both give identical results.
All parameters in experiments use their default values unless otherwise specified.
The codebase implements multiple languages. To run the BFF variant from Section~\ref{sec:bff}, pass \texttt{-{}-lang bff\_noheads}.

\subsection{Long tape simulations}

An alternative environment that we explored consists of a single long tape (usually 65,536 bytes) where all the code and data reside. These are what we call ``Long tape simulations''. Here, at each iteration, we select a random initial position in the tape and execute instructions sequentially until a set limit of operations is achieved or if an error occurs.
In this kind of simulations, since there are no ``individual strings'' that could determine an atomic self-replicator, it becomes even more apparent that a self-replicator is a substring. 

We confirmed that BFF variants also generate self-replicators in long tape settings. One key difference with the primordial soup simulations is that we need to decide what to do about \texttt{head0} and \texttt{head1}'s initial positions. If those are set the same as in the initial PC, we find that trivial (non-looping) self-replicators rapidly take over the universe. Adding an offset to head1 is sufficient to allow looping self-replicators to arise. Anecdotally, we appear to require an offset somewhat larger than 8 to work (e.g., 12 or 16). While we do not discuss this experiment in detail, we release its code at \url{https://github.com/benlaurie/bff-ben/tree/paper2}, and the experiment can be reproduced using command \texttt{MAXPROCS=32 go run -tags graphics links.org/bf/cmd/bfsoup}. In Section~\ref{forth_long_tape} we explore long tapes in much more detail for a different programming language.

\section{Rise of Self-replicators with other languages}
While we do not yet have a general theory to determine what makes a language and environment amenable to the rise of self-replicators, we have observed this behavior in settings other than BFF. In this section, we provide further evidence of self-replication appearing on computational substrates that are significantly different from BFF: Forth -- a stack-based protocol for computation, and emulations of a real-world Z80 CPU architecture. We also show a \textit{counterexample} of a language where, despite significant effort, we did not observe self-replicators: \texttt{SUBLEQ}.

\subsection{Forth}\label{sub:forthsoup}

Forth refers more to a family of languages than a specific one.
In general, Forth variants differ from BFF by having a stack.
Input commands from the instruction tape either push values onto the stack or
perform operations on the stack. In particular, this allows the stack to take
the same role as the heads in BFF programs, in that it controls where instructions
read and write from in the tape.
More details can be found on Wikipedia~\cite{enwiki:1215242468}.

We present two variants of Forth for two different settings: the tape-pairing setting we've previously termed ``primordial soup'', and a ``long tape'' setting where individual Forth interpreters execute in parallel on different parts of the tape, but there is no concept of individual, competing, tapes.

Despite self-replicators emerging in both of these settings, we didn't find an instruction set that works ``out-of-the-box'' for both settings.

\subsubsection{Primordial soup simulations}\label{primordial_forth}

Inspired by BFF, we use a variant of Forth with a restricted instruction set, as well as fixed-size tapes that interact with each other through concatenation. The particular instruction set we use dedicates half of the instruction space for jump instructions, and another quarter for stack push instructions. The complete instruction set is as follows, where \texttt{<top>} refers to the value at the top of the stack, \texttt{<top - n>} refers to the n-th value below the top of the stack, \texttt{<pc>} is the instruction pointer and \texttt{push/pop} add or remove a value from the stack, respectively.

\begin{center}
\begin{tabular}{>{\tt}cc>{\tt}l}
0000 0000 && <top> = *<top> \\
0000 0001 && <top> = *(<top> + 64) \\
0000 0010 && *<top> = <top - 1>; pop; pop \\
0000 0011 && *(<top> + 64) = <top - 1>; pop; pop \\
0000 0100 && push <top> \\
0000 0101 && pop \\
0000 0110 && swap <top - 1> and <top> \\
0000 0111 && if <top> != 0: <pc>++ \\
0000 1000 && <top> = <top> + 1 \\
0000 1001 && <top> = <top> - 1 \\
0000 1010 && <top - 1> = <top> + <top - 1>; pop \\
0000 1011 && <top - 1> = <top> - <top - 1>; pop \\
0000 1100 && *(<top> + 64) = *<top>; pop \\
0000 1101 && *<top> = *(<top> + 64); pop \\
01xx xxxx && push [xxxxxx] \rm{(unsigned)} \\
1Xxx xxxx && <pc> = <pc> $\pm$ ([xxxxxx]+1)\textrm{, sign depends on} X \\
\end{tabular}
\end{center}

An interesting property of this Forth variant is that it admits a trivial one-byte self-replicator: executing \texttt{0C} on an empty stack will copy itself over onto the first byte of the other string. The programs exhibit quite interesting dynamics: a self-replicator that copies whole tapes appears fairly quickly, building upon the existing one-byte self-replicator to achieve its functionality. 

The following is an example of a ``complete'' self-replicator that emerges in this setup:

\newlength{\linecenter}
\newlength{\overlinewidth}
\newcommand{\forthinsn}[3][2em]{%
  \setlength{\linecenter}{\dimexpr(#1/2)\relax}%
  \settowidth{\overlinewidth}{\text{\tt\footnotesize{#3}}}%
  \overset{%
    \parbox[c]{#1}{%
      \rotatebox[origin=l]{90}{%
        \parbox[c][\linecenter]{\overlinewidth}{%
          \text{\tt\footnotesize{#3}}%
        }
      }%
    }%
  }{%
    \parbox[c][1em]{#1}{%
      \text{\tt #2}%
    }
  }
}

\[
\forthinsn{0C}{COPY+64}%
\forthinsn{08}{INC}%
\forthinsn{04}{DUP}%
\forthinsn{81}{JUMP +2}%
\forthinsn{2C}{NOP}%
\forthinsn{C4}{JUMP -5}%
\]

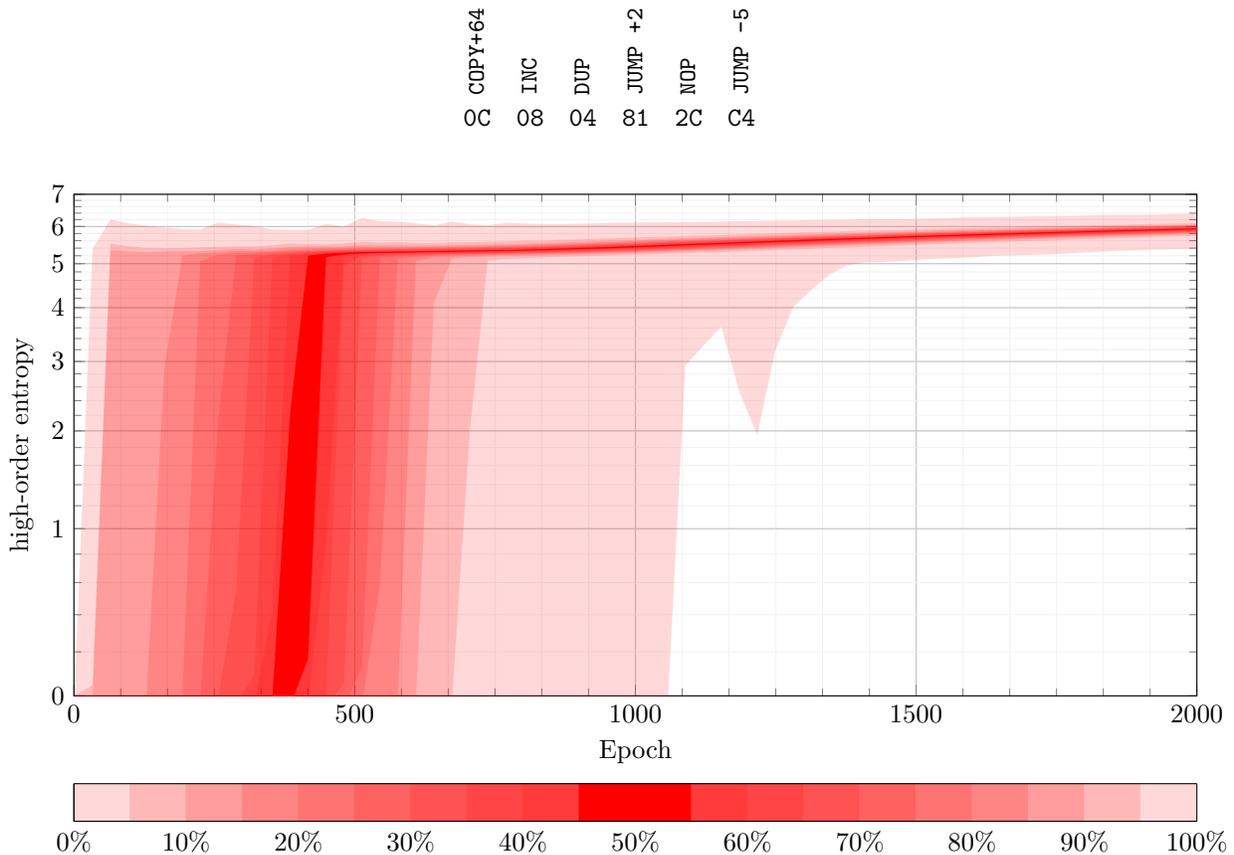
\begin{figure}[h]
  
\begin{tikzpicture}
\newcommand{\defquantile}[1] {
\addplot[draw=none, forget plot, name path=perc#1] table[col sep=comma, header=false, x index=0, y expr=1+\thisrowno{#1}] {forthtrivial_runstats.csv};
}
\newcommand{\shadequantile}[3] {
\addplot [fill=red, opacity=#3, forget plot] fill between[of=perc#1 and perc#2];
}
\newcommand{\quantilelegend}[2]{
\addlegendimage{area legend, fill=red, opacity=#1, draw=none}\addlegendentry{$#2$}
}
\begin{axis}[
    xlabel=Epoch,
    ylabel=high-order entropy,
    width=\textwidth,
    height=0.5\textwidth,
    colorbar horizontal,
    colorbar style={
        at={(0,-0.25)},
        anchor=south west,
        xticklabel = {$\pgfmathparse{\tick*5}\pgfmathprintnumber[fixed,precision=1]{\pgfmathresult}$\%},
        xtick style={draw=none},
    },
    colormap name=shadesofred,
    colorbar as palette,
    x tick label style={
        /pgf/number format/.cd,
            fixed,
            1000 sep={},
            fixed zerofill,
            precision=0,
        /tikz/.cd
    },
    ymode=log,
    ytick={1,...,8},
    minor ytick = {1,1.2,...,8},
    yticklabel = {
        \pgfmathparse{e^\tick-1}
        $\pgfmathprintnumber[fixed,precision=1]{\pgfmathresult}$
    },
    xtick={0,500,...,2000},
    grid=both,
    minor tick num=5,
    grid style={line width=.1pt, draw=gray!10},
    major grid style={line width=.2pt,draw=gray!50},
    xmin=0,
    ymin=1,
    ymax=8,
    mark size=0,
    xmax=2000,
]
\defquantile{1}
\defquantile{2}
\defquantile{3}
\defquantile{4}
\defquantile{5}
\defquantile{6}
\defquantile{7}
\defquantile{8}
\defquantile{9}
\defquantile{10}
\defquantile{11}
\defquantile{12}
\defquantile{13}
\defquantile{14}
\defquantile{15}
\defquantile{16}
\defquantile{17}
\defquantile{18}
\defquantile{19}
\defquantile{20}
\defquantile{21}

\shadequantile{1}{21}{0.15}
\shadequantile{2}{20}{0.15}
\shadequantile{3}{19}{0.15}
\shadequantile{4}{18}{0.15}
\shadequantile{5}{17}{0.15}
\shadequantile{6}{16}{0.15}
\shadequantile{7}{15}{0.15}
\shadequantile{8}{14}{0.15}
\shadequantile{9}{13}{0.15}
\shadequantile{10}{12}{1.0}

\end{axis}
\end{tikzpicture}
\caption{Distribution of complexity for Forth programs in the primordial soup setting over time across 1000 different runs, with $0.024\%$ mutation rate. Each shade of red represents a different quantile range. Almost all runs show a state transition within 1k epochs.}
\label{fig:forth_complexity_analysis}
\end{figure}

Figure~\ref{fig:forth_complexity_analysis} shows evolution of complexity over time for this Forth variant. Comparing with Figure~\ref{fig:complexity_analysis},
we can see that self-replicators emerge much more consistently and quickly, as could be expected by the relative simplicity of self-replicator.

\begin{figure}[t]
    \centering
    \begin{subfigure}[t]{0.24\textwidth}
        \includegraphics[width=\textwidth]{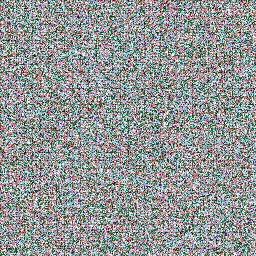}
    \end{subfigure}%
    \hfill%
    \begin{subfigure}[t]{0.24\textwidth}
        \includegraphics[width=\textwidth]{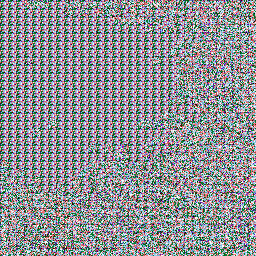}
    \end{subfigure}
    \hfill
    \begin{subfigure}[b]{0.24\textwidth}
        \includegraphics[width=\textwidth]{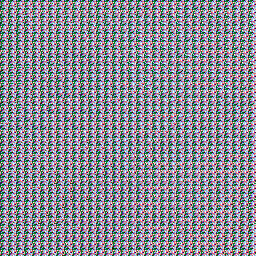}
    \end{subfigure}%
    \hfill%
    \begin{subfigure}[b]{0.24\textwidth}
        \includegraphics[width=\textwidth]{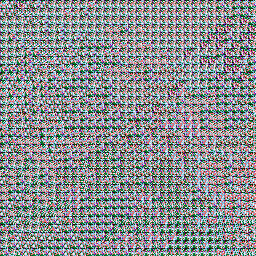}
    \end{subfigure}
    \caption{Visualization of a 2d Forth primordial soup, in order from left to right: before state transition, at the start of it,
    just after the majority of programs has become a self-replicator, and many evolution steps afterwards. A full video of this evolution can be found at \url{https://www.youtube.com/watch?v=eOHGBuZCswA}.
    }
    \label{fig:forth2d}
\end{figure}

\paragraph{2D primordial soup simulations}
This Forth formulation produces self-replicators in higher dimensions as well: Figure~\ref{fig:forth2d} shows some snapshots
of a Forth run on a 2D grid. From random programs (far left), at first one self-replicator arises (center left) and quickly takes over the entire space (center right). Later, we observe a long-term differentiation of the programs in the soup (far right).

Here are three self-replicators that can be found in the soup after 20 thousand evolution steps:

\[
\forthinsn{0C}{COPY+64}%
\forthinsn{08}{INC}%
\forthinsn{3F}{NOP}%
\forthinsn{15}{NOP}%
\forthinsn{04}{DUP}%
\forthinsn{92}{JUMP +19}%
\forthinsn{...}{}%
\forthinsn{D7}{JUMP -24}%
\hspace{4cm}
\forthinsn{0C}{COPY+64}%
\forthinsn{09}{DEC}%
\forthinsn{1C}{NOP}%
\forthinsn{27}{NOP}%
\forthinsn{04}{DUP}%
\forthinsn{92}{JUMP +19}%
\forthinsn{...}{}%
\forthinsn{D7}{JUMP -24}%
\]
\[
\forthinsn{0C}{COPY+64}%
\forthinsn{09}{DEC}%
\forthinsn{1F}{NOP}%
\forthinsn{1F}{NOP}%
\forthinsn{04}{DUP}%
\forthinsn{C4}{JUMP -5}%
\]

These self-replicators are slightly different from one another and show how the soup is never dominated by one individual program, but instead competition remains present even after a long time.

The code for the experiments above can be found at \url{https://github.com/paradigms-of-intelligence/cubff}.
To run the Forth variant, pass \texttt{-{}-lang forthtrivial}.

\subsubsection{Long tape simulations}\label{forth_long_tape}

The long tape setting -- an environment where there is only one long, contiguous ``program'' or tape -- admits several Forth variants that produce replicators. In our experiments we have (for performance reasons) used multiple threads running simultaneously. We sacrifice determinacy for performance by foregoing locking (which also means that all counters are approximate). We do not believe that this materially affects behaviour.

We investigated using the same instruction set as in Section~\ref{primordial_forth}, modifying tape-pair-specific instructions to operate either at an offset against the current PC, or offset against a ``pseudo-tape'' marker positioned at the nearest preceding 64-byte aligned position in the long tape to the PC. Neither of these variants gave rise to replicators, however their viability was verified by running an experiment where the long-tape was seeded with a hand-crafted replicator, consisting of:

\[
\forthinsn{40}{PUSH 0}%
\forthinsn{40}{PUSH 0}%
\forthinsn{40}{PUSH 0}%
\forthinsn{00}{READ 0}%
\forthinsn{03}{WRITE 1}%
\forthinsn{08}{INC}%
\forthinsn{04}{DUP}%
\forthinsn{04}{DUP}%
\forthinsn{C7}{JMP -7}%
\]

This replicator uses the stack to copy itself, by iteratively copying one value to the stack, then writing it to the ``second pseudo tape" (a simple offset by 64), and repeating this indefinitely. When seeded with this replicator, it proceeded to occupy the entire long tape, further complexify, and persist indefinitely.

The main variant we chose to investigate in more detail, which exhibits the spontaneous inception of replicators, uses the following instruction set:

\begin{center}
\begin{tabular}{>{\tt}cc>{\tt}l}
0000 xxxx && push [xxxx]{\rm~(sign-extended)} \\
0001 xxxx && <top> = (<top> << 4) + [xxxx] \\
0010 0000 && *(<pc> + <top - 1> + <top>) = *(<pc> + <top - 1>); pop 1 \\
0010 0001 && inc <top> \\
0010 0010 && dec <top> \\
0010 0011 && jump to <pc> + <top - 1> if <top> != 0, pop 2 \\
\end{tabular}
\end{center}

All other bit patterns are no-ops. With a single tape of length 65,536 bytes and a rate of 1 random mutation to a new valid instruction per 400,000 instructions executed, we generally see ``good'' replicators emerge in approximately 60 seconds, or 180 billion instructions. In our experiments, we achieve around $3 \cdot 10^9$ instructions per second across all threads.

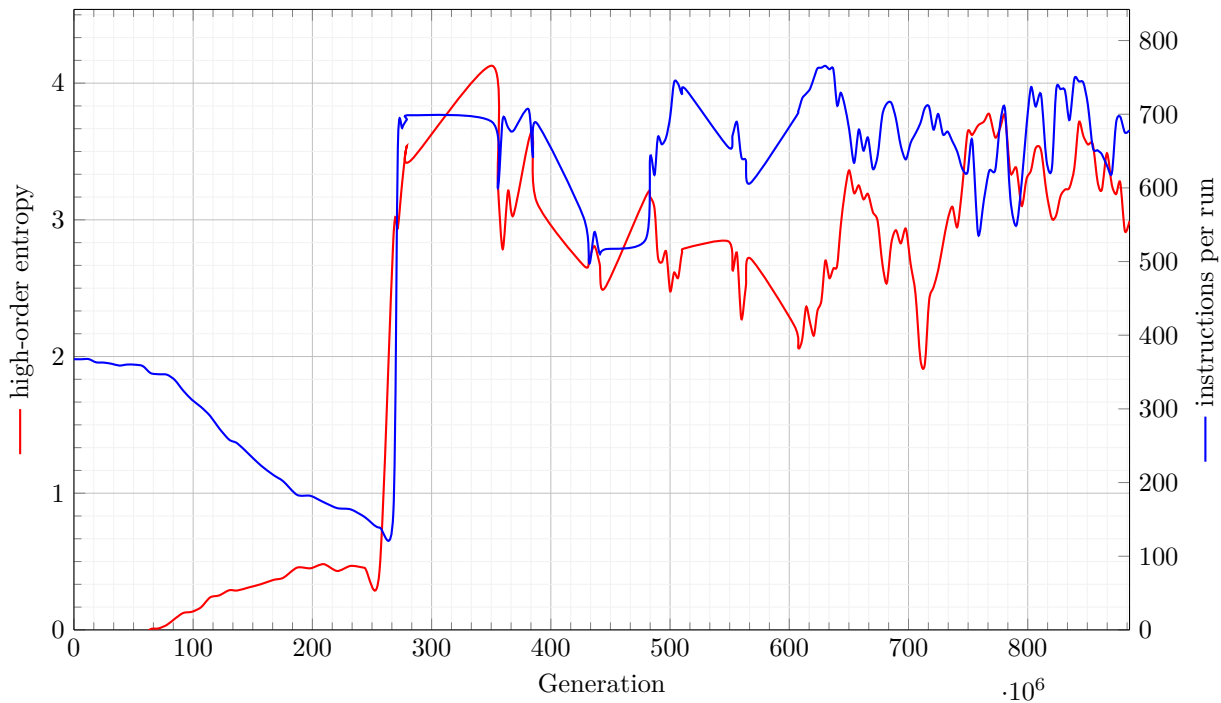
\begin{figure}
\begin{tikzpicture}
\pgfplotsset{set layers}
\begin{axis}[
    scale only axis,
    axis y line*=left,
    xlabel=Generation,
    ylabel={\ref{complexityplot} high-order entropy},
    width=0.85\textwidth,
    height=0.5\textwidth,
    x tick label style={
        /pgf/number format/.cd,
            fixed,
            1000 sep={},
            fixed zerofill,
            precision=0,
        /tikz/.cd                                               
    },
    scaled x ticks = base 10:-6,
    grid=both,
    minor tick num=5,
    grid style={line width=.1pt, draw=gray!10},
    major grid style={line width=.2pt,draw=gray!50},
    xmin=0, xmax=885399084,
    ymin=0,
    mark size=0,
]
\addplot[smooth, thick, color=red] table[col sep=comma, x=generation, y=entropy-cratio] {f5run.csv}; \label{complexityplot}
\end{axis}
\begin{axis}[
    scale only axis,
    axis y line*=right,
    axis x line=none,
    width=0.85\textwidth,
    height=0.5\textwidth,
    ylabel={\ref{rateplot} instructions per run},
    xmin=0, xmax=885399084,
    ymin=0,
    mark size=0,
]
\addplot[smooth, thick, color=blue] table[col sep=comma, x=generation, y=rate] {f5run.csv}; \label{rateplot}
\end{axis}
\end{tikzpicture}
\caption{Evolution of complexity over time for a long-tape Forth run.}
\label{f5}
\end{figure}

Each thread chooses a random PC and executes from there until an error occurs (stack overflow/underflow are the only possible errors), or a fixed number of instructions have been executed (1,000 in this example). This process is then repeated. We use 8 threads in this example.

Figure~\ref{f5} shows the evolution of complexity over time for a single long-tape Forth run. ``Bad'' replicators (ones that only take over part of the universe) arise within the first few seconds. These generally do not loop but instead consist only of a series of PUSHes and COPYs which tend to populate the locality with more of the same, but do not, in practice, extend further before mutations prevent them from working correctly. This corresponds to the initial increase of high-order entropy observed in Figure~\ref{f5}. As mentioned above, ``good'' replicators emerge in around a minute or so, corresponding to the abrupt rise in high-order entropy shown in Figure~\ref{f5}. Note the corresponding increase in average instructions executed per run. A typical ``good'' replicator looks something like this:

\resizebox{\linewidth}{!}{
$
\forthinsn{09}{PUSH \phantom{+}-7}%
\forthinsn{0B}{PUSH \phantom{+}-5}%
\forthinsn{0C}{PUSH \phantom{+}-4}%
\forthinsn{03}{PUSH \phantom{++}3}%
\forthinsn{0E}{PUSH \phantom{+}-2}%
\forthinsn{01}{PUSH \phantom{++}1}%
\forthinsn{01}{PUSH \phantom{++}1}%
\forthinsn{09}{PUSH \phantom{+}-7}%
\forthinsn{09}{PUSH \phantom{+}-7}%
\forthinsn{09}{PUSH \phantom{+}-7}%
\forthinsn{01}{PUSH \phantom{++}1}%
\forthinsn{01}{PUSH \phantom{++}1}%
\forthinsn{09}{PUSH \phantom{+}-7}%
\forthinsn{09}{PUSH \phantom{+}-7}%
\forthinsn{09}{PUSH \phantom{+}-7}%
\forthinsn{01}{PUSH \phantom{++}1}%
\forthinsn{09}{PUSH \phantom{+}-7}%
\forthinsn{09}{PUSH \phantom{+}-7}%
\forthinsn{01}{PUSH \phantom{++}1}%
\forthinsn{01}{PUSH \phantom{++}1}%
\forthinsn{10}{SHIFT \phantom{+}0}%
\forthinsn{09}{PUSH \phantom{+}-7}%
\forthinsn{01}{PUSH \phantom{++}1}%
\forthinsn{1D}{SHIFT -3}%
\forthinsn{20}{COPY}%
\forthinsn{21}{INC}%
\forthinsn{09}{PUSH \phantom{+}-7}%
\forthinsn{01}{PUSH \phantom{++}1}%
\forthinsn{23}{JNZ}%
$
}

An interesting feature we observe in the emergent replicators is that they tend to consist of a fairly long non-functional head followed by a relatively short functional replicating tail. The explanation for this is likely that beginning to execute partway through a replicator will generally lead to an error, so adding non-functional code before the replicator decreases the probability of that occurrence. It also decreases the number of copies that can be made and hence the efficiency of the replicator, resulting in a trade-off between the two pressures. In the replicator above, the functional tail involves the last 7 instructions, starting at \verb+PUSH 1+. Note that this loop actually copies the head of the ``next'' replicator rather than its own head.

As in the BFF experiments, we also see replicators change over time, though they tend to remain broadly similar to the example above. Nevertheless, sometimes we find very short replicators, which tend to be very stable, unlike the long ones. For example:

\[
\forthinsn{0D}{PUSH -3}%
\forthinsn{07}{PUSH \phantom{+}7}%
\forthinsn{20}{COPY}%
\forthinsn{21}{INC}%
\forthinsn{0A}{PUSH -6}%
\forthinsn{0E}{PUSH -2}%
\forthinsn{23}{JNZ}%
\]

The only instruction that changed over time was the one before last (\texttt{PUSH -2}) which can push any non-zero value without affecting functionality. Note that the loop omits the first instruction, which would otherwise lead to a stack overflow.

The code for this experiment is available at \url{https://github.com/benlaurie/bff-ben/tree/paper1}, using command line \texttt{GOMAXPROCS=32 go run --tags="graphics" links.org/bf/cmd/f5}.

\subsection{SUBLEQ}\label{sub:subleq}
We also experimented in primordial soup simulations with \texttt{SUBLEQ}, one of the simplest Turing-complete languages, possessing a single instruction.
In \texttt{SUBLEQ}, there is only one piece of state -- the program counter \texttt{pc}. Executing an instruction consists
in reading values \texttt{a}, \texttt{b} and \texttt{c}, starting at the program counter (\texttt{pc}). The instruction that is executed
then is (in C-like syntax):

\begin{center}
\texttt{*a -= *b; if (*a <= 0) \{ goto c; \} else \{ goto pc + 3; \}}
\end{center}

The smallest hand-crafted self-replicator we managed to write with \texttt{SUBLEQ} is $60$ bytes. This may be too long and hint at some length requirements for self-replicators to arise. To investigate further, we constructed a different \texttt{SUBLEQ} variant, which we call \texttt{RSUBLEQ4}, which admits a significantly shorter self-replicator.

In this variant, each instruction reads $4$ values (\texttt{a}, \texttt{b}, \texttt{c}, \texttt{d}), and then executes:

\begin{center}
\texttt{*(pc + a) = *(pc + b) - *(pc + c); if (*a <= 0) \{ goto pc + d; \} else \{ goto pc + 4; \}}
\end{center}

The following is a 25 byte self-replicator in \texttt{RSUBLEQ4}\footnote{\url{https://asciinema.org/a/oHvCby3FKzSOoZ835Bl8BzEJl}}:

\begin{center}
\texttt{9 16 20 4 4 5 19 4 0 0 12 4 -3 -3 9 4 -8 8 -7 -12 0 -1 -1 -64 -73}
\end{center}

In both variants, the program terminates when the counter moves to a position that would require reading an out-of-bounds value. For both variants, we confirmed that if we seeded the soup with one self-replicator, self-replicators would quickly and often take over the entire soup.

Nevertheless, when randomly initialized, the soup remained in almost complete random uniformity even following billions of executions. There are no dynamics to change the distribution of strings, and self-replicators are too rare to be generated by random background mutation. We note that the first self-replicators observed to arise in other substrates all have much shorter lengths than the ones we hypothesized possible in \texttt{SUBLEQ} variants. We believe that this counterexample could be a valuable starting point for constructing a theory that predicts what languages and environments could harbor life, perhaps by modeling the likelihood of the simplest self-replicator to arise based on variables proportional to the length of such self-replicator.

The code for the experiments above can be found at \url{https://github.com/paradigms-of-intelligence/cubff}.
To run the \texttt{SUBLEQ} and \texttt{RSUBLEQ4} variants, pass \texttt{-{}-lang subleq} and \texttt{-{}-lang rsubleq4} respectively.

\subsection{Real-world instruction sets}

Previous sections explore spontaneous emergence of self-replicators and state transition phenomena in a few computational substrates based on artificially designed minimalistic languages. In order to test the generality of our observations we perform an experiment with a system using emulation\footnote{\url{https://github.com/superzazu/z80}} of the real-world Z80 CPU architecture \cite{carr1980z80}. We study a 2D grid of 16-byte programs initialized with uniform random noise. At each simulation step we randomly pick a pair of adjacent tapes ``A'' and ``B'' and concatenate them in random order (``AB'' or ``BA''). Then we reset the Z80 emulator and run 256 instruction steps, using the concatenated tapes a the memory buffer. All memory read and write request addresses are computed by modulo of the concatenated tape length (32 bytes), which prevents out-of-bounds accesses by random programs. In parallel to CPU-driven self-modification process, mutations are applied to random bytes of the grid.

\begin{figure}
{\includegraphics[width=\textwidth]{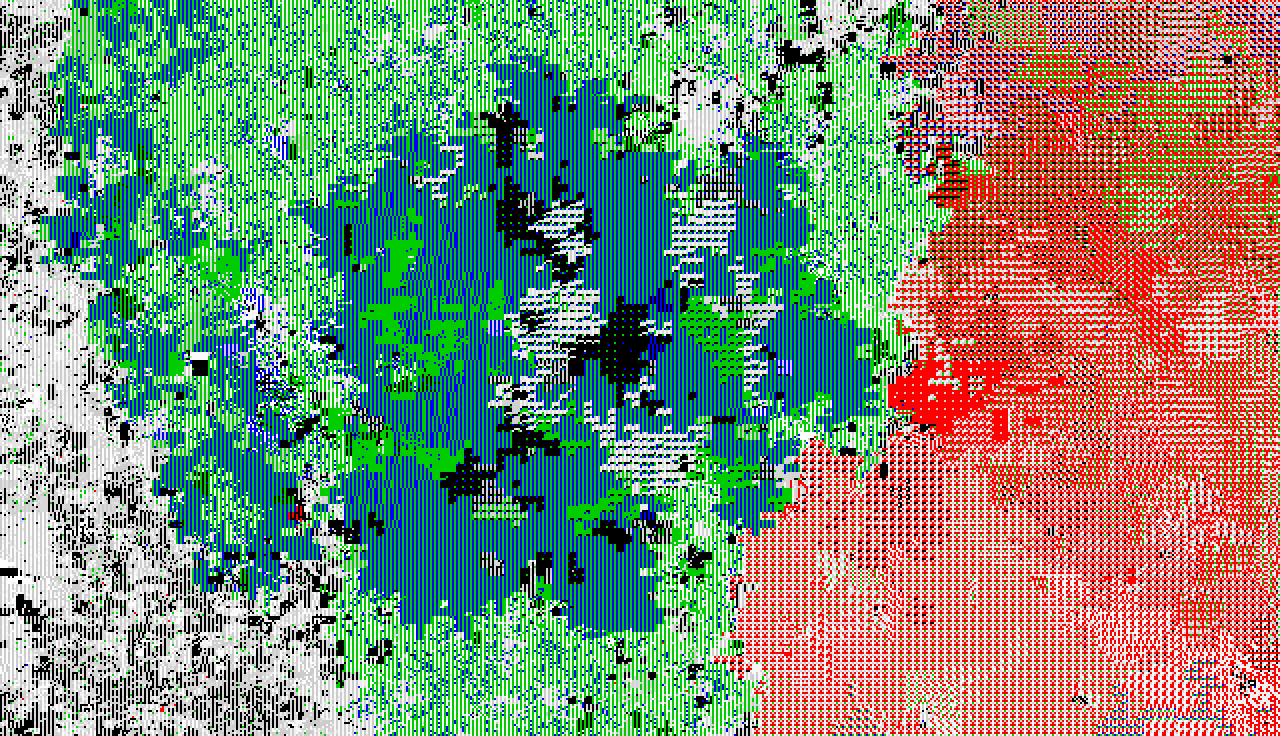}};
\caption{Ecosystem of self-replicators produced by Z80 CPUs operating on a 2D grid. Every 4x4 group of pixels correspond to a single 16-byte program. At every simulation step a random pair of adjacent cells gets selected, concatenated and executed by a Z80 emulator for 256 steps. We observe emergence of a few generations of self-replicators. First the wave of stack based self-replicators sweeps the grid and forms an ``ecosystem'' of a few co-existing variants. Then the grid is overtaken by more robust self-replicators that use memory copy instructions. Colors correspond to a few most popular instruction codes used by self-replicators: {\color{red}\texttt{LDIR/LDDR}} - memory copy, {\color{blue} \texttt{PUSH HL}} - push 16-bits (stored in H and L registers) onto stack, {\color{ForestGreen} \texttt{LD HL,X / LD HL,(X)}} - set HL registers with immediate or indirect value.}
\label{fig:z80}
\end{figure}

This simple setup gives rise to surprisingly complex behaviours with a number of self-replicator generations exploiting different Z80 features emerging. Some of these self-replicators form a sort of symbiotic ecosystems, while other compete for domination (Figure~\ref{fig:z80}). We often observe a series of state transition-like events when more and more capable self-replicators or replicator collectives overtake the soup multiple times. Early generations use stack-based copy mechanism: at initialization Z80 sets the stack pointer at the end of the address space, so pushing values onto stack gives tape A a simple mechanism of writing to tape B. Most of the time we see the development of an ``ecosystem'' of stack-based self-replicators that eventually gets replaced with self-replicators that exploit ``LDIR'' or ``LDDR'' instructions that allow to copy continuous chunks of memory. We created an interactive lab to facilitate the exploration of z80 self-replicators: \url{https://github.com/znah/zff}.

We have also tried the 8080 CPU in the long-tape setting. This produces replicators which seem to always be two bytes repeated, for example, 01 c5 -- which, if execution starts on the 01, corresponds to LXI BC, 01c5 (01 c5 01), PUSH BC (c5) -- which has the effect of setting the top of the stack to 01 c5. If execution starts on c5, then BC will be 0 for the first push, and so 00 00 will be written to memory. 00 in 8080 is a no-op, so this is harmless. Note that these replicators are non-looping, which, in long-tape BFF, are not able to take over all of memory. However, these replicators work very well in 8080. Perhaps for this reason we have never seen a looping variant emerge in 8080.

\section{Discussion}

In this paper we showed examples of several computational substrates where life---identified by the rise and dominant take-over by self-replicating programs---can emerge from a pre-life period. We showed how variants of BF spontaneously create self-replicators from primordial soups of different dimensionalities mostly due to self-modification with or without background mutation rates. We also showed anecdotal evidence that this is the beginning of more complex dynamics. We showed how different languages and paradigms such as Forth and Z80 and 8080 CPUs also result in similar behaviors. Finally, we showed a counterexample with \texttt{SUBLEQ}-like languages where we were unable to catalyze the spontaneous emergence of self-replicators. In our preliminary analyses, \texttt{SUBLEQ}-like languages seem to have a much higher expected length for a functioning initial self-replicator. We believe that such a length plays a critical role in determining how likely self-replicators are to arise but we expect it to not be the sole factor at play.

We argue that this set of computational substrates shows a new way of discovering and arriving at life. The behavior of such systems is markedly different from auto-catalytic networks and biologically-inspired systems. Our analysis starts at the pre-life period as opposed to the experiments performed in Tierra and AVIDA where they began with hand-crafted self-replicators. Unlike previous work on computational substrates focused on pre-life where they observed self-replicators arise due to random initialization or mutation~\cite{RASMUSSEN1990111, rasmussen1991matter, PARGELLIS1996111}, we showed that self-modification is the main culprit for self-replicator to arise in most of the experiments we performed. Moreover, our initial explorations and the ones observed in similar systems such as Tierra, AVIDA and Coreworld suggest that this may be just the beginning of the complexity of behaviors that can emerge and flourish in such systems.

Several open questions arise from these investigations that warrant further investigations. How much complexity can spontaneously arise in open-ended computational systems? What are the distinguishing properties of a system that encourages or inhibits the rise of self-replicators? Is there a way for us to guide the evolution of such systems into developing increasingly complex functions? Finally, what kind of evolution can arise in these computational systems? Would it be similar to what we observe in nature or would it manifest notable differences?

We look forward to exploring more of these ideas and hope that this will bring us closer to understanding the limits and potential of life, irrespective of the substrate on which it emerges.

\section*{Acknowledgments}
We thank Thomas Fischbacher, João Sacramento, Alexander Meulemans and Stan Kerstjens for their thoughtful feedback.

\bibliographystyle{unsrt}  
\bibliography{main}

\end{document}